\documentclass[preprint,12pt]{elsarticle}
\usepackage{amssymb}
\usepackage{setspace} % 导入控制行距的宏包

\journal{Pattern Recognition} % 指定目标期刊名称

\usepackage{caption}
\usepackage{multirow}
\usepackage[table,xcdraw]{xcolor}
\usepackage{hyperref}

\usepackage{natbib}
\usepackage{ragged2e}

\usepackage{lmodern} % for better font rendering

\usepackage{bm}

\usepackage{algorithm}
\usepackage{algpseudocode}
\usepackage{amsmath}
\usepackage{bm}
\usepackage{rotating} % 用于支持横置

\usepackage{wrapfig}

\hypersetup{
	colorlinks=True,
	citecolor=cyan,
	linkcolor=cyan,
	urlcolor=cyan
}
\usepackage{setspace}
\begin{document}
	\begin{frontmatter}
		\title{Desensitizing for Improving Corruption Robustness in Point Cloud Classification through Adversarial Training}

		\author[label1]{Zhiqiang Tian}
		\author[label1]{Weigang Li\corref{cor1}}
		\cortext[cor1]{Corresponding author}
		\ead{liweigang.luck@foxmail.com}
		\author[label2]{Chunhua Deng}
		\author[label1]{Junwei Hu}
		\author[label1]{Yongqiang Wang}
		\author[label3]{Wenping Liu}
		
		\address[label1]{School of Information Science and Engineering, Wuhan University of Science and Technology, Wuhan 430081, China}
		\address[label2]{School of Computer Science and Technology, Wuhan University of Science and Technology, Wuhan, 430065 , China}
		\address[label3]{School of Information Management and Institute of Big Data and Digital Economy, Hubei University of Economics, Wuhan, China}

		\begin{abstract}
			Due to scene complexity, sensor inaccuracies, and processing imprecision, point cloud corruption is inevitable. Over-reliance on input features is the root cause of DNN vulnerabilities. It remains unclear whether this issue exists in 3D tasks involving point clouds and whether reducing dependence on these features can enhance the model's robustness to corrupted point clouds. This study attempts to answer these questions. Specifically, we quantified the sensitivity of the DNN to point cloud features using Shapley values and found that models trained using traditional methods exhibited high sensitivity values for certain features. Furthermore, under an equal pruning ratio, prioritizing the pruning of highly sensitive features causes more severe damage to model performance than random pruning. We propose `Desensitized Adversarial Training' (DesenAT), generating adversarial samples using feature desensitization and conducting training within a self-distillation framework, which aims to alleviate DNN's over-reliance on point clouds features by smoothing sensitivity. First, data points with high contribution components are eliminated, and spatial transformation is used to simulate corruption scenes, generate adversarial samples, and conduct adversarial training on the model. Next, to compensate for information loss in adversarial samples, we use the self-distillation method to transfer knowledge from clean samples to adversarial samples, and perform adversarial training in a distillation manner.Extensive experiments on ModelNet-C and PointCloud-C demonstrate show that the propose method can effectively improve the robustness of the model without reducing the performance of clean data sets. This code is publicly available at \href{https://github.com/JerkyT/DesenAT/tree/master}{https://github.com/JerkyT/DesenAT}.
		\end{abstract}
	
	\end{frontmatter}

\section{Introduction}
The main goal of this study is to enhance the robustness of the 3D point cloud classification model to data corruption by reducing the model's over-reliance on specific input features. In complex real-world scenarios, point cloud data inevitably suffer from corruption owing to factors such as viewpoint changes, environmental conditions, and acquisition devices \cite{huang2024joint, shen2021interpreting, PointNeXt}. The emergence of large-scale corrupted point cloud datasets, such as ModelNet40-C \cite{ModelNet-C} and Pointcloud-C \cite{Pointcloud-C}, highlights the importance of addressing these challenges. Current mainstream approaches improve data generalization through techniques such as fusion \cite{Pointcutmix, RSMix} and interpolation \cite{Pointmixup}, but often overlook the inherent vulnerability of deep neural networks (DNNs) to corrupted point clouds. In particular, DNNs tend to rely excessively on certain feature points, making them vulnerable when dealing with corrupt data.

Before proposing this method, it is essential to explore the effect of corrupted point clouds on DNNs. We aimed to quantify the contribution of each feature component to model predictions, representing the model's sensitivity to various point cloud features, and to study how corruption transformations affect DNN robustness. This study used Shapley values to compute feature attribution because Shapley values, from a game-theoretic perspective, are the only unbiased estimators of the input variable attribution \cite{sundararajan2020many}.

\begin{figure}{l}
	\centering
	\includegraphics[width=0.7\linewidth]{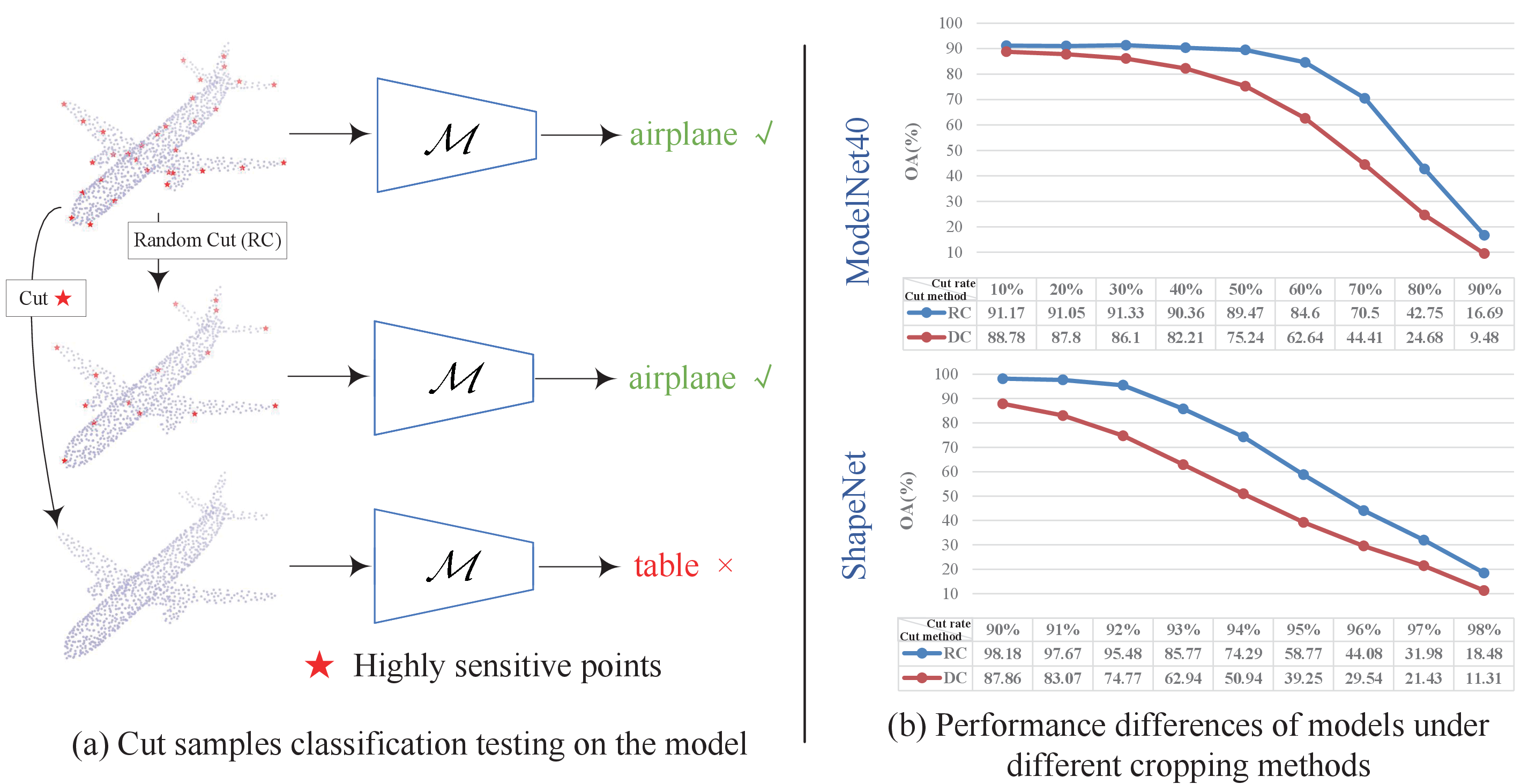}
	\caption{(a) Visualizes Shapley values, showing that the DNN is sensitive to certain features in clean point clouds. However, after corruption transformations, these features undergo changes. The neural network demonstrates recognition capability for both the original and randomly pruned samples. When high-sensitivity data features are pruned, the model fails to correctly classify the sample. (b) We conducted experiments on ModelNet-C and ShapeNet, as the pruning ratio increases, the recognition capability of the DNN gradually decreases. Under the same pruning ratio, prioritizing the pruning of high-sensitivity features is more threatening compared to a random pruning strategy.}
	\label{Fig1}
\end{figure}

We adopted a general approach to address the vulnerability of the models to corrupted point clouds rather than focusing on specific types of corruption. As shown in Fig. \ref{Fig1} (a), we first used Shapley values to quantify the contribution of each feature in the sample to the model's prediction and identify specific high-contribution features. By applying corruption through feature pruning, we observed that removing high-contribution features (Contribution Pruning, CP) poses a greater threat to DNN performance than random pruning (Random Pruning, RP). In Fig. \ref{Fig1} (b), the table shows that under the same pruning ratio, the model finds it more difficult to recognize the CP data than the RP data. We hypothesized that DNNs in 3D classification tasks are overly reliant on certain key features, and the loss or alteration of these points due to corruption (e.g., pruning) leads to decreased robustness.

To explore this further, we compared the feature contribution distribution of DNNs trained via Standard Training (ST) and Adversarial Training (AT). This study shows that models trained with AT (which enhances robustness compared to ST) exhibit a smoother distribution of feature contributions, suggesting that adversarially trained neural networks are less dependent on specific features.

Inspired by this, we propose a universal `Desensitizing Adversarial Training' (DesenAT) method. This method reduces in the dependence of DNNs on point clouds features by smoothing the sensitivity of DNNs to features. Unlike most existing AT methods, the proposed approach does not target a specific corruption type but addresses robustness issues from a universal perspective. Specifically, we developed a model that employs the ST method. Using Shapley values, the contribution of each feature in the sample to the model was quantified. We then removed these high-contribution features to mitigate their impact on the DNN and introduce random spatial transformations to disrupt the original form of the adversarial samples. Next, we retraind the network by employing the generated adversarial samples. On one hand, we generated adversarial samples within permissible limits to maximize loss, even if this confuses the network model. On the other hand, while ensuring significant sample variations, we aimd for the neural network model to have a sufficiently small loss on the training data, providing the model with robustness to adapt to such data transformations. Finally, to compensate for the information loss in adversarial samples, we employed knowledge distillation to transfer knowledge from clean samples to adversarial samples.

Extensive experiments on ModelNet-C and PointCloud-C demonstrate that the proposed approach effectively enhances the robustness of the model without compromising its performance on clean data. Moreover, the proposed method does not introduce additional computational burden. This approach exhibits strong performance across common corruption types compared to training methods specifically tailored for certain feature corruption variations. The versatility of this method is a notable advantageous, because it demonstrates good efficacy across a range of prevalent corruption types. The contributions of this study are as follows:

1. We have posited a novel hypothesis in this study: Corruption transformations can alter or even disrupt the feature representation of point clouds, leading to reduced robustness in models overly reliant on these features.

2. We propose the `Desensitizing Adversarial Training' method to reduce the excessive reliance of DNNs on certain features, thereby improving robustness. Compared to existing training methods, the proposed approach is more universal.

3. We employed knowledge distillation to facilitate information transfer from clean to adversarial samples, thus mitigating information loss. This method bridges the gap in information between clean and adversarial samples.

\section{Related Work}\label{s2}
\subsection{Robustness in Point Cloud}\label{s2.1}
Liu et al. \cite{liu2019extending} exposed point clouds' vulnerability to adversarial attacks, laying the foundation for corruption resistance studies. ModelNet-C and PointCloud-C were introduced as standardized benchmarks to assess the robustness of point cloud models against common corruptions.

Several augmentation methods have since been developed to boost generalization and robustness. RSMix \cite{RSMix} randomly samples and mixes points from different point clouds, reducing model dependency on specific configurations, while PointMixup \cite{Pointmixup} uses linear interpolation between point clouds to combat density corruption. Pointcutmix \cite{Pointcutmix} enhances generalization by cutting and mixing object sections, introducing structural diversity.

More recent strategies focus on improving robustness. CausalPC \cite{huang2024causalpc} identifies causal effects in classification, while \cite{lou2024hide} generates imperceptible but rational adversarial perturbations. PointCert \cite{zhang2023pointcert} provides deterministic certified robustness, and CAP \cite{ding2023cap} enhances classification via semantic and structural modeling.

Although these methods introduce geometric variations that improve generalization, they do not fully address the over-reliance of deep neural networks (DNNs) on specific features, leading to vulnerability under adversarial conditions. Our method promotes a more comprehensive understanding of object geometry by smoothing feature sensitivity, thereby increasing robustness across corruption types.

\subsection{self Knowledge Distillation}\label{s2.2}
With the rapid progress of deep learning \cite{liu2022automated, krawczyk2023segmentation}, data-driven algorithms \cite{sun2023weighted, yang2014adaptive, hao2023contrastive} have become a dominant paradigm. Knowledge distillation, using strategies like mutual and self-learning, allows neural networks to utilize unlabeled and cross-modal data effectively, making it a key area in deep learning research \cite{qi2023contrast}. Self-distillation methods are widely used in semi-supervised and unsupervised tasks by generating soft labels from the data itself to guide learning. For example, Shen et al. \cite{shen2022self} proposed Self-Distillation from Last Mini-Batch (DLB), which achieves consistency regularization by learning from the previous mini-batch, improving model performance and robustness. Kim et al. \cite{kim2021self} introduced Progressive Self-Knowledge Distillation (PSKD), which iteratively refines model-generated targets to enhance learning. Unlike these approaches, our method eliminates inter-batch information interaction and focuses on transferring information from original to adversarial samples within the same batch, offering a unique mechanism for knowledge distillation.

\subsection{Application of Shapley Value in Neural Network Model}\label{s2.3}
From a game theory perspective, the Shapley value is widely used as the only unbiased off the input variable attribution. Lundberg et al. \cite{lundberg2017unified} first applied the Shapley value to deep networks and demonstrated its rationality in explaining the deep network process. Zheng et al. \cite{zheng2022shap} proposed an interpretable framework for CNNs called shap-CAM, which uses the Shapley value to determine the importance of each pixel, thereby eliminating the dependency on gradients in the CAM algorithm. Wang et al. \cite{wang2021shapley} embedded the Shapley value into deep models as an inner layer, providing new perspectives and tools for the interpretability of neural networks. Teneggi et al. \cite{teneggi2022fast} proposed a hierarchical Shapley (h-Shap), an image classification model-agnostic explanation method based on the hierarchical extension of Shapley coefficients, addressing some of the limitations of the current methods. Shen et al. \cite{shen2021interpreting} introduced the Shapley value to 3D point cloud classification tasks for the first time and, concluded that neural network models are more sensitive to the boundaries and corners of point clouds based on neural network sensitivity to point cloud regions. 

These methods aforementioned methods all employ the Shapley value for interpretability tasks, focusing on understanding the decision-making process of the model. Conversely, this study not only provides a general explanation for the vulnerability of neural network models through the Shapley value—neural networks are overly sensitive to certain feature points, leading to their vulnerability—but also introduces the Shapley value into the training of 3D point clouds for the first time, thereby designing an adversarial training framework that effectively enhances the robustness of the model.

\section{Corruption-Robust Point Cloud Classification via Adversarial Training with Shapley Value}\label{s3}

%\begin{figure}[!h]
%	\centering
%	\includegraphics[width=1\linewidth]{LCT.eps}
%	\caption{Schematic diagram of the framework DesenAT. First, we assess the model's sensitivity to the original samples using Shapley values, then divide the point cloud into high- and low-sensitivity points. By removing high-sensitivity points, we reduce the model's reliance on them. We generate adversarial samples through spatial transformations and define a distillation loss based on feature differences between original and adversarial samples. The model is then trained to minimize this loss, achieving knowledge transfer.}
%	\label{LCT}
%\end{figure}

This section introduces our adversarial training approach for 3D point cloud models using Shapley value analysis. First, we explain the Shapley values (Sec. \ref{s3.0}) and then describe the adversarial training process. In Sec. \ref{s3.1}, Shapley values are used to quantify the contribution of point cloud features to DNN predictions. We found that corruption can cause changes or loss of key features, thereby reducing the model robustness. To address this, Sec. \ref{s3.2} proposes an adversarial training method that removes high-contributing components to generate adversarial samples, reducing the model's feature dependence. In Sec. \ref{s3.3}, knowledge distillation is used to transfer knowledge from clean to corrupt samples, thereby mitigating information loss.

\subsection{Preliminary Knowledge}\label{s3.0}
\subsubsection{Shapley Value in Game Theory}\label{s3.0.1}
	The shapley value \cite{grabisch1999axiomatic} is a concept from game theory used to allocate the total payoff in cooperative games to the various participants. It is calculated based on the contributions of each participant to the cooperation. For any cooperative game involving \(N\) participants, the contribution of participant \(i\) can be represented as follows:
	\begin{equation}\label{1}
		\begin{array}{l}
			{\phi _i}(v) = \frac{1}{{N!}}\sum\limits_{|S| < N,i \in S}{\frac{{(N - 1)! \cdot (|S| - 1) \cdot (N - |S|)!}}{{N!}}}\cdot (v(S) - v(S\backslash \{i\})),
		\end{array}
	\end{equation}
	where \(S\) represents a subset of \(N\), indicating the cooperative group, \(v\) represents the value function, and \(v(S)\) represents the total payoff from the cooperation of all members within the subset \(S\). \(|S|\) denotes the number of members in the subset \(S\), and \(S \cup \{i\}\) represents the new group formed by adding member \(i\) to set \(S\).

\subsubsection{Adversarial Training}\label{s3.0.2}
	The adversarial training process is typically described as follows: Sample $X$ and its label $y$ both follow the underlying distribution $D$. The loss function is defined as $L(X,y,\theta)$, where $\theta$ represents the model weights. Adversarial training can be defined as follows:
	\begin{equation}\label{2}
		\mathop {{\rm{argmin}}}\limits_\theta {{\rm E}_{(X,y) \sim D}}[\mathop {\rm{max}}\limits_{||\delta || \le \varepsilon }L(X + \delta, y, \theta)],
	\end{equation}
	where $\delta$ represents noise. The expression above constitutes the a process of maximization and minimization. The goal of maximization is to find an adversarial sample that maximizes the loss function, whereas the goal of minimization is to find the weights $\theta$ that minimizes the adversarial loss.

\subsection{Quantification of Shapley Value Distribution of Point Cloud}\label{s3.1}
To investigate the impact of each feature on DNN predictions, we employed Shapley values to quantify the contribution of each feature component. Specifically, a point cloud sample $X$ consists of $N$ points, and the neural network is denoted as ${\cal M}$. We treat each point in $X$ as a member, the output probability of the target class as the gain, and according to Eq. \ref{1}, we can use the Shapley value to represent the contribution of the $i$-th point to the target class:
\begin{equation}\label{3}
\begin{array}{l}
	{\phi _i}= \frac{1}{{N!}}\sum\limits_{S \subseteq X,i \in S}{\frac{{(N - 1)! \cdot (|S| - 1) \cdot (N - |S|)!}}{{N!}}}\cdot 
	(\tau (S,{\cal M}) - \tau (S\backslash \{i\},{\cal M})) ,
\end{array}
\end{equation}
where $S$ represents a subset of $X$, and $\tau (S,{\cal M})$ represents the score of the sample's target class after subset $S$ is passed through the neural network ${\cal M}$ to extract features. By employing Eq. \ref{3}, we can obtain the contribution of each point in $X$. The contribution matrix can be represented as follows: $\bm{\Phi}= \{{\phi _1},{\phi _2},{\phi _3},...,{\phi _i},...,{\phi _N}\}$.

Considering the permutation invariance of the point clouds, it is not possible to compare the differences and similarities in the Shapley values between different samples by comparing the corresponding points. Because the distribution of data is independent of its order, this study analyzes the impact of different training methods on robustness of the model from the Shapley distribution perspective.

Specifically, we define the number of bins as $b$, with each bin having a length of $1/b$. The elements in $\Phi$ are allocated to the corresponding bins based on their values, and their quantities are counted. The distribution matrix $\hat \Phi$ of $\Phi$ can be represented as follows:
\begin{equation}\label{4}
	\begin{array}{l}
	\hat \Phi = {D_H}(\Phi ) = \{{d_1},{d_2},...,{d_i},...{d_b}\},
	\end{array}
\end{equation}
where ${D_H}(\cdot)$ represents the histogram-based distribution function, and ${d_i}$ represents the number of elements in the $i$-th bin. The visualization results are shown in Fig. \ref{Fig2} (e) and (f).

\subsubsection{Quantitative Analysis of Single Samples}\label{s3.1.2}
\begin{figure}{l}
	\centering
	\includegraphics[width=0.7\linewidth]{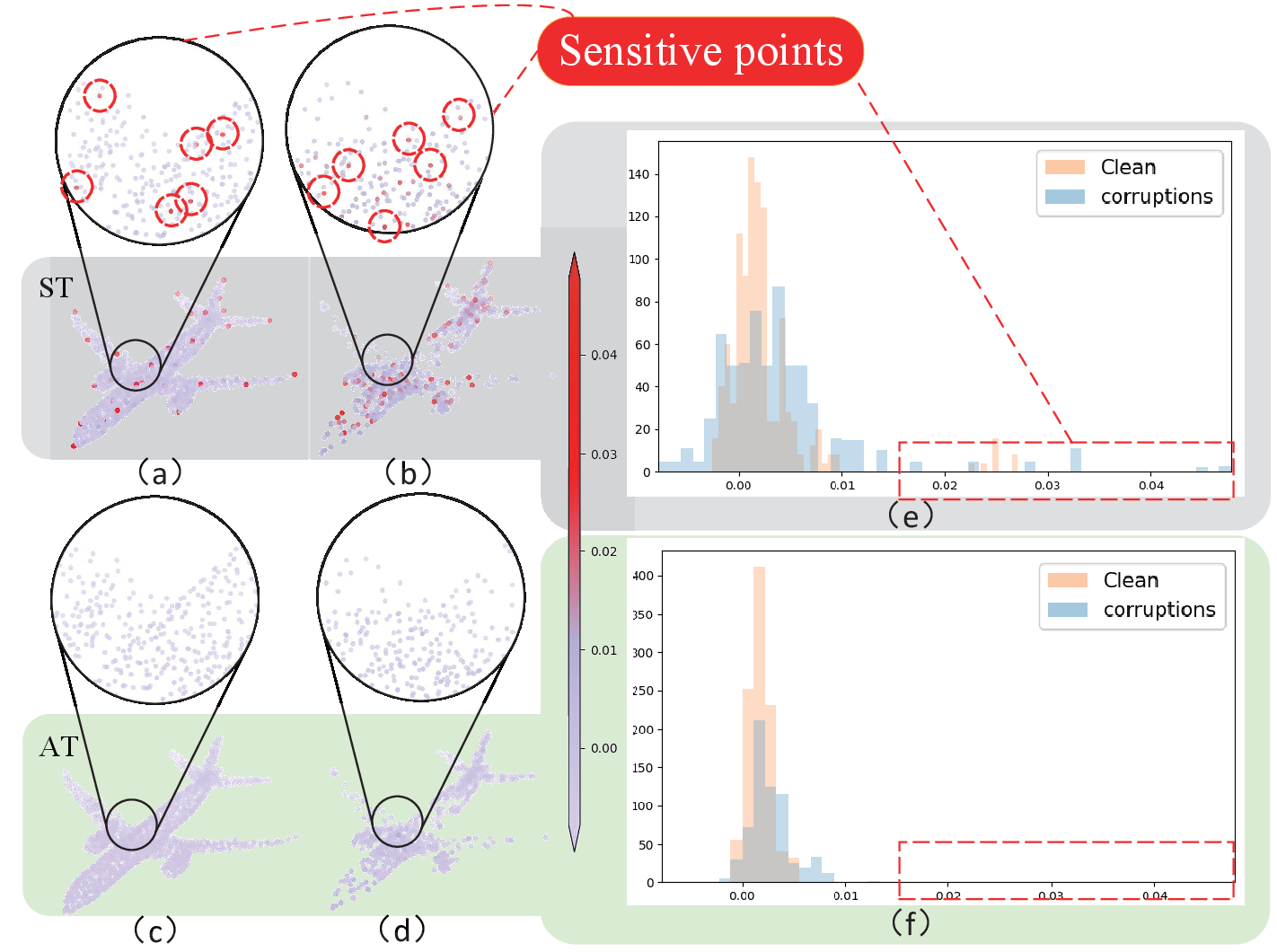}
	\caption{Provides visualizations of Shapley values for both clean and corrupted point clouds obtained employing the ST method with PointNet++(MSG) as the learning model, as shown in panels (a) and (b). Similarly, corresponding visualizations of Shapley values for clean and corrupted point clouds obtained employing the AT method with PointNet++(MSG) as the learning model are presented in panels (c) and (d). Additionally, histograms in panels (e) and (f) depict the corresponding distributions of Shapley values.}
	\label{Fig2}
\end{figure}

As shown in Fig. \ref{Fig2}, we employed the Standard Training (ST) method and the Adversarial Training (AT) methods to train the PointNet++(MSG) model to evaluate their robustness (AT $>$ ST). We then computed the contribution of each point in the test samples to the target class for both clean and corrupted test samples. 

In Fig. \ref{Fig2}, the color of the points in the point cloud samples (a), (b), (c), and (d) represent the model's sensitivity to them. Points closer to purple indicate lower sensitivity, whereas points closer to red indicate higher sensitivity. From (a) and (b), it can be seen that the model trained with the ST method has high sensitivity values for a few key points in the point cloud samples. This suggests that the model trained with the ST method determines the class of the point cloud by focusing on a few `key points', which leads to its vulnerability: when corruption changes alter these key points or there is a misidentification of the `key points' (as shown in (b)), the model becomes overconfident and makes misclassifications. In contrast, (c) and (d) show that the model trained with the AT method has a smoother sensitivity distribution across the feature points in the point cloud samples, without `high sensitivity points.' This indicates that the model trained with the AT method classifies the sample based on its overall features rather than focusing on specific points, making the model more robust: even if corruption changes some feature points, the model can still distinguish the class using global information.

\subsubsection{Global Quantitative Analysis}\label{s3.1.3}
%\mbox{}\par
\begin{figure}{l}
	\centering
	\includegraphics[width=0.7\linewidth]{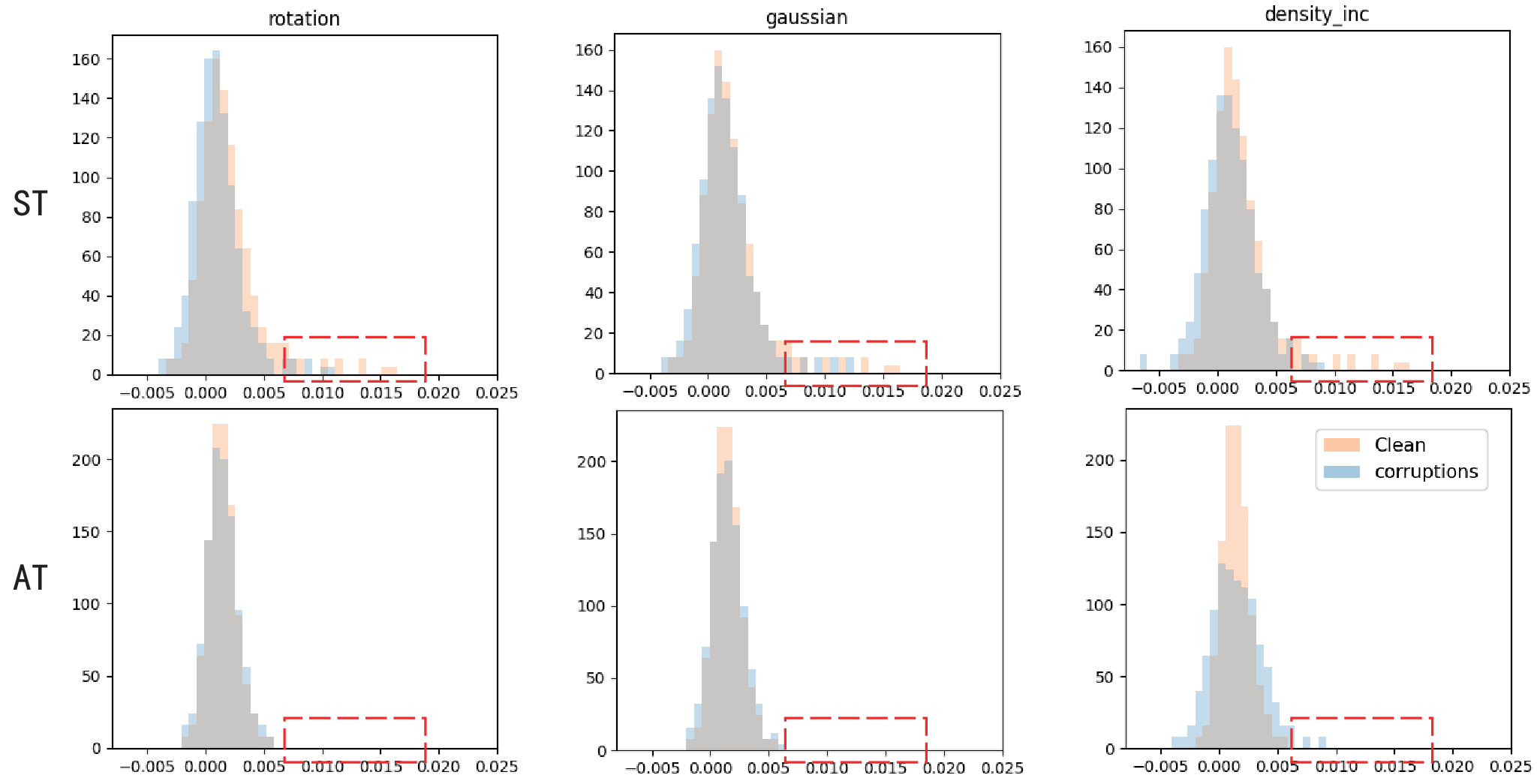}
	\caption{Visualization of the contribution value distribution of samples to PointNet++(MSG) in the ModelNet40-C dataset (3 benchmarks: rotation, gaussian, density, corruption level is 5.) under different training methods. `ST' denotes standard training and `AT' denotes adversarial training. It is worth noting that in adversarial training, we employed data augmentation methods similar to those of corrupt types to generate adversarial samples for targeted training.}
	\label{Fig3}
\end{figure}
The phenomenon in Sec. \ref{s3.1.2} was observed on individual samples. To validate its generality, we analyzed the entire dataset. Specifically, we define the entire dataset as $\textbf{X}$ or $\textbf{X}= \{{X_1},{X_2},...,{X_i},..,{X_\Theta}\}$, where $\Theta$ is the number of samples. By combining Eqs. \ref{3} and \ref{4}, we obtain the distribution off each sample:
\begin{equation}\label{5}
	\begin{array}{l}
		\bm{\hat \Phi}= \{{\hat \Phi _1},{\hat \Phi _2},...,{\hat \Phi _i},...,{\hat \Phi _\Theta}\},
	\end{array}
\end{equation}
where $\bm{\hat \Phi}$ is a matrix in dimensions $\Theta \times N$ where $N$ is the number of points in each sample. We sorted all the points in each sample according to their Shapley values, and calculate the average Shapley value for the $\Theta$ points in the same order. This gives the following new distribution:
\begin{equation}\label{6}
	\begin{array}{l}
		\left\{{\begin{array}{*{20}{c}}
				{\bm{\hat \Phi}' = \{{{\bar d}_1},{{\bar d}_2},...,{{\bar d}_i},...,{{\bar d}_N}\}}\\
				{{{\bar d}_i}= \mathrm{mean}({{\hat \Phi }_i})}\\
		\end{array}}\right.,
	\end{array}
\end{equation}
where $\bm{\hat \Phi}'$ has a dimension of $N$, and ${\hat \Phi }_i$ represents the $i$-th sample. We consider it as the contribution value distribution of the entire dataset.

We employed the entire ModelNet40 and ModelNet40-C test sets for validation. Using Eq. \ref{6}, we computed the overall contribution value distributions for both the clean and corrupted datasets using different training methods, as illustrated in Fig. \ref{Fig3}.

%It can be intuitively seen that there are some data points with high shapley in the red box area of `ST', while there are almost no such points in the red box area of `AT', which indicates that the model under ST method relies on key data points to judge the category, while the model under AT method can use the information from all data points in a balanced manner. This phenomenon is the same as the rule in Sec. \ref{s3.1.2}, and its universality can be verified. Based on the above observations, we conclude that the neural network model relies too much on key data points in the classification task, which makes it not robust to the corruption point cloud. Our goal is to see if we can generate adversarial samples by shapley to induce a more uniform sensitivity of the network to sample midpoints.

The red box in `ST' contains high-Shapley data points, while `AT' lacks such points, showing that the ST model relies on key points for classification, whereas the AT model uses all points more evenly. This aligns with the rule in Sec. \ref{s3.1.2}, confirming its universality. Thus, neural networks overly depend on key points, reducing robustness to corrupted point clouds. Our goal is to generate adversarial samples using Shapley to achieve more uniform network sensitivity to sample midpoints.

\subsection{Adversarial Training of 3D Point Cloud}\label{s3.2}
\begin{figure}{l}
	\centering
	\includegraphics[width=0.6\linewidth]{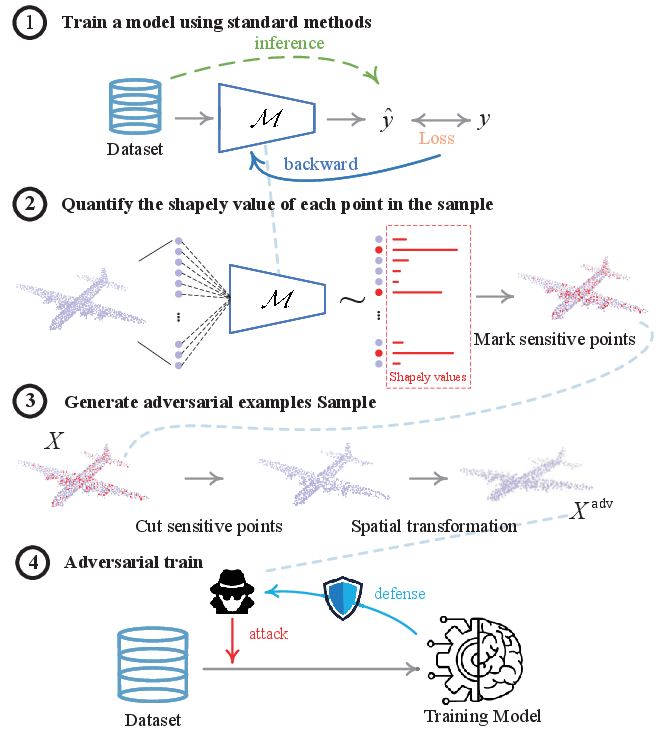}
	\caption{Adversarial training process diagram. where $\hat{y}$ denotes the predicted category of the model.}
	\label{Fig4}
\end{figure}

Building on the insights gleaned from the aforementioned, the proposed approach to mitigate the network's sensitivity to key points involves removing highly sensitive points during the training process. This compels the model to acquire knowledge from points with lower sensitivity, thereby inducing it to make overall judgments about the target categories in classification tasks. In particular, as shown in Fig. \ref{Fig4}.

\textcircled{1}. We first employ the ST method to train a baseline model $\cal M$;

\textcircled{2}. Using Eq. \ref{3}, we compute the Shapley values for all data points in the training samples. Points with higher Shapley values contribute more significantly to the target class of the sample, indicating a higher sensitivity of the neural network model.

\textcircled{3}. By excluding these highly-contributing data points, we reduce the model's sensitivity to the samples, compelling the model to learn knowledge from all data points evenly. Notably, when removing these highly sensitive points, we introduce a random ratio $r$. This ratio is consistent for all samples within the same batch during training, thereby ensuring dimensional consistency for batch processing. However, this ratio may vary among the batches. This approach enhances the sample generalization, enabling the neural network to adapt to inputs with different numbers of points.

In addition to filtering point cloud samples and reducing the sensitivity to data, we propose a universal spatial transformation method. Thsi aims to corrupt the adversarial samples. Specifically, unlike 2D images, 3D point cloud data is more `three-dimensional', and the data points are unordered. Simple morphological changes do not hinder human recognition. In 2D images, as shown in Eq. (2), an adversarial sample can be simply defined as $X + \delta$ without any spatial changes. However, owing the disorder and 3D point clouds, spatial transformations are indispensable for generating adversarial samples. In this study, the spatial transformation of 3D point clouds is defined follows:
\begin{equation}\label{7}
		{X^{\rm adv}}= k \cdot X + \delta,
\end{equation}
where $k$ denotes a random transformation matrix of (3 $\times$ 3) and $\delta$ denotes random noise.

\textcircled{4}. After generating adversarial samples, we redefine the position of the data transformation in Eq. \ref{2}. The expression for adversarial training in point cloud tasks is as follows:
\begin{equation}\label{8}
	\left\{
	\begin{array}{c}
		\displaystyle \min_{\theta}\mathbb{E}_{(x,y)\sim D}\left[ \max_{X^{\rm{adv}} \cong (X)} L(f_{\theta}({X^{\rm{adv}}}),y) \right] \\
		{X^{\rm adv}} = k \cdot F(X,r) + \delta,
	\end{array}
	\right.,
\end{equation}
where $F(\cdot)$ is the Shapley-based filtering function, $r$ represents the randomly chosen filtering ratio, and ${X^{\rm{adv}}} \cong X$ indicates that ${X^{\rm{adv}}}$ can vary within the range permitted by $X$. On the one hand, within the allowable range, we aim to find that maximizes the loss between the model's predicted probability and the true label, even if it means generating adversarial samples that attempt to `perceive' the model as much as possible. On the other hand, while generating adversarial samples that are suitably corrupted, we seek to obtain the network weights $\theta$ such that the expectation of the target class for the sample is minimized as much as possible. We enhanced the robustness of the model using this adversarial approach.

\subsection{Self-Distillation Adversarial Training}\label{s3.3}

\begin{figure}{l}
	\centering
	\includegraphics[width=\linewidth]{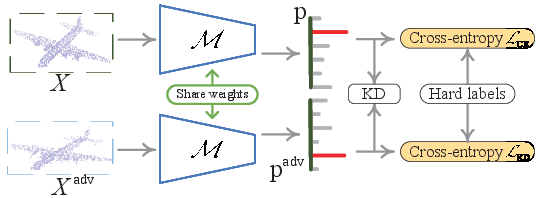}
	\caption{Adversarial self-distillation framework.}
	\label{Fig5}
\end{figure}

This above method re-trains the network by removing data points, thereby reducing the model sensitivity. However, when removing the data points, we lost some useful information. Consequently, we propose a complementary form of DesenAT called `desensitizing adversarial training-self distillation' (DesenAT-sD). Building on the adversarial training in Sec. \ref{s3.2}, this framework introduces clean samples as input, effectively compensating for the information loss in adversarial samples.

The forward propagation of the network in Fig. \ref{Fig5} can be expressed as follows: $\mathbf{p},\mathbf{p}^{\rm{adv}}= {\cal M}(X,{X^{\rm{adv}}})$, where ${\mathbf{p}}=[{p_1},{p_2},...,{p_i},...,{p_c}] \in {^{1 \times c}}$, $c$ is the total number of classes, $p_i$ is the probability of class $i$ in the classification task, and the score corresponding to the true label can be represented as ${\mathbf{q}}= [{q_1},{q_2},...,{q_i},...,{q_c}] \in R{^{1 \times c}}$. If $y$ is the index of the true class and $k$ is any class index, then for all cases where $k \ne y$, ${q_y}= 1$ and ${q_k}= 0$, where $X$ and $X^{\rm{adv}}$ represent different features of the same sample, therefore, their probability abstracted through ${\cal M}$ should satisfy ${\mathbf{p}}\equiv {\mathbf{p}}{^{\rm{adv}}}$.

We adopted the distillation approach to facilitate knowledge transfer between $\mathbf{p}$ and $\mathbf{p}^{\rm{adv}}$. For the predicted probabilities $\mathbf{p}$ from the network output, considering any two classes $i$ and $j$ with probabilities $p_i$ and $p_j$, their mutual information can be described employing joint probability: $V_{ij}= {p_i}{p_j}$. In this study, inspired by the cross-entropy formula ${\rm{H(}}i{\rm{)}}= {q_i}\log {p_i}$, we define the joint entropy as: ${\rm{H(}}i,j{\rm{)}}= {q_i}{q_j}\log {p_i}{p_j}$, and employ a fully connected graph to depict the interrelationships among all classes:
\begin{equation}\label{9}
\left\{{\begin{array}{*{20}{c}}
		{{\mathbf{A = }}{{({\mathbf{p}})}^{\rm{T}}}{\mathbf{p}}}\\
		{{{\mathbf{A}}^{\rm{adv}}}{\mathbf{= }}{{({\mathbf{p}})}^{\rm{T}}}{{\mathbf{p}}^{\rm{adv}}}}\\
\end{array}}\right.,
\end{equation}
where $\mathbf{A}$ denotes a fully connected graph, where the position $\mathbf{A}_{i, j}$ can be interpreted as the mutual information between categories $i$ and $j$.

Subsequently, we employs graph matrices to represent inter-class entropy:
\begin{equation}\label{10}
	{{\mathbf{A}}^{\rm{H}}}{\mathbf{= A}}\log {\mathbf{(}}{{\mathbf{A}}^{\rm{adv}}}{\mathbf{)}}.
\end{equation}

The distilled loss for an individual sample can be defined as follows:
\begin{equation}\label{11}
	{{\cal L}_{KD}}{\rm{(}}{\mathbf{A}}\parallel {{\mathbf{A}}^{\rm{adv}}}{\rm{) = - }}\sum\limits_{i = 1}^c {\sum\limits_{j = 1}^c {{\mathbf{A}}_{ij}^{\rm{H}}}},
\end{equation}
simultaneously, we must calculate the loss between the network output probabilities $\mathbf{p}$, $\mathbf{p}^{\rm{adv}}$ and the hard label $\mathbf{q}$:
\begin{equation}\label{12}
	{{\cal L}_{\rm{CE}}}= {{\cal L}_{\rm{CE}}}(\mathbf{p},\mathbf{q}) + {{\cal L}_{\rm{CE}}}(\mathbf{p^{\rm{adv}}},\mathbf{q}),
\end{equation}
by combining Eqs. \ref{11} and \ref{12}, the loss function can be represented as follows:
\begin{equation}\label{13}
	{\cal L}{\rm{= }}\alpha \cdot {{\cal L}_{KD}}+ (2 - \alpha) \cdot {{\cal L}_{\rm{CE}}},
\end{equation}
where $\alpha$ is the weighting coefficients for the loss.

\section{Experiments and Analysis}\label{s4}

This chapter comprises three sections. In Sec. \ref{s4.1}, we describe the experimental setup. Sec. \ref{s4.2} presents comparative experiments to evaluate the state-of-the-art performance of the proposed algorithm. In Sec. \ref{s4.3}, we present ablation experiments to assess the effectiveness of each branch in the innovative approach.

\subsection{Setup}\label{s4.1}
\subsubsection{Dataset}\label{s4.1.1}
\textbf{ModelNet40-C \cite{ModelNet-C}:} This is a comprehensive dataset designed to test the robustness of corruption in 3D point cloud classification tasks. \cite{ModelNet-C} created a dataset based on the ModelNet test set encompassing three modes of corruption: transformation, density and noise. Each corruption mode includes five common corruption types (denoted by $C = 5$) in the point clouds: transformation \{Rotation, Shear, FFD, RBF, Inv RBF\}, density \{Occlusion, LiDAR, Local Density Inc, Local Density Dec, Cutout\}, and noise \{Uniform, Gaussian, Impulse, Upsampling, Background\}. Each type of corruption had five severity levels ($S = 5$), resulting in 75 benchmarks(15 $\times$ 5). For each benchmark, the number of samples is 2468 (the same as the number of samples in the test set of the ModelNet dataset), with 1024 points per sample (except for the `cutout' and `background' benchmarks). The total number of samples in the entire dataset was 185,100 (2468 $\times$ 75), enabling the evaluation of a model's robustness across different corruption types.

\textbf{PointCloud-C \cite{Pointcloud-C}:} Similar to ModelNet40-C, PointCloud-C is a testing benchmark for conducting robustness analysis of 3D point clouds in corrupted scenarios based on the ModelNet40 \cite{ModelNet40} test set. This dataset simulates real-world point cloud corruptions from three perspectives: `object', `sensor', and `processing' \{Scale, Rotate, Jitter, Drop, Global, Add Global \}. It encompasses seven corruption types, each having five severity levels, resulting in 35 benchmarks in total.

\subsubsection{Networks}\label{s4.1.2}
This approach focuses on training strategies and is independent of network architecture. Four 3D vision models were evaluated: PointNet++(MSG) \cite{Pointnet++}, PointNetMeta-S \cite{pointNetMetaBase2023cvpr}, and two transformer-based models, APES\_global and APES\_Local \cite{APES2023cvpr}.

\subsubsection{Metric}\label{s4.1.3}
We utilized the mean Corruption Error (mCE) proposed in \cite{ModelNet-C} by employing the PointNet++(MSG) model obtained using the ST method as the baseline to standardize the corruption resistance of different models and training methods. To calculate mCE, we first present the formula for Corruption Error(CE):
	\begin{equation}\label{15}
	{{\rm{CE}}_i}= \frac{{\sum\nolimits_{l = 1}^S {(1 - {{\rm{OA}}_{i,l}})}}}{{\sum\nolimits_{l = 1}^S {(1 - {\rm{OA}}_{i,l}^{PointNet + + (MSG,ST)})}}},
	\end{equation}
where ${\rm{OA}}_{i,l}^{PointNet++(MSG,ST)}$ denotes the overall accuracy (OA) of PointNet++(MSG) on the test set of corruption-type $i$ at the corruption level $l$ within 1,2,..,$S$, mCE represents the mean CE across multiple corruption types:
	\begin{equation}\label{16}
	{\rm{mCE}}= \frac{1}{C}\sum\limits_{i = 1}^C {{{\rm{CE}}_i}},
	\end{equation}
Furthermore, we provided the performances of various models and methods under the mean overall accuracy (OA) metric:
	\begin{equation}\label{17}
	{\rm{mOA}}= \frac{1}{S}\sum\limits_{i = 1}^S {{{\rm{OA}}_i}},
	\end{equation}
where $l$ represents the corruption level. In both the ModelNet40-C and PointCloud-C datasets, the corruption level was set to $S$ = 5.

\subsubsection{Training Details}\label{s4.1.4}
All experiments were conducted on a Tesla V100S-PCIE-32GB GPU using the PyTorch framework. The input channels $(x, y, z)$ were normalized to [0, 1] during training, and no data augmentation was applied in standard training. Hyperparameters were consistent with the original paper and related references \cite{ModelNet-C, Pointcloud-C}. For PointNet++(MSG) and PointNetMeta-S, settings followed \cite{pointNetMetaBase2023cvpr}, while APES\_global and APES\_Local used \cite{APES2023cvpr}. In adversarial training, the random range for sensitivity point exclusion rate $r$ (Sec. \ref{s3.3}) was [0.5, 1]; for transformation matrix $k$ (Eqs. \ref{7}, \ref{8}), [0, 0.25]; and for noise $\delta$, [0, 0.03]. The weight coefficient $\alpha$ in Eq. \ref{13} was set to 1.

Notably, the training process in this study was solely conducted on clean datasets. In other words for any given method, training occureds only once on clean data sets, and the obtained model was subsequently tested on various corruption types and benchmarks.

\subsection{Comparative Experiment}\label{s4.2}

% 15 OA
\begin{table}[!h]
	\centering
	\caption{The mOA(\%) of different model architectures on ModelNet40-C with different training strategies. Bold: best in column. Underline: second best in column. \dag: single-scale grouping (SSG) version, \ddag: multi-scale inference from \cite{RepSurf}, ST: standard training. Red: Above ST, Green: Below ST.}\label{tab 1}
	\resizebox{\linewidth}{!}{
		\begin{tabular}{llrrrrrrrrrrrrrrrr} \hline
			&                               & \multicolumn{1}{c}{\textbf{Rotation}}      & \multicolumn{1}{c}{\textbf{Shear}}         & \multicolumn{1}{c}{\textbf{FFD}}           & \multicolumn{1}{c}{\textbf{RBF}}           & \multicolumn{1}{c}{\textbf{Lnv.RBF}}       & \multicolumn{1}{c}{\textbf{Uniform}}        & \multicolumn{1}{c}{\textbf{Ups.}}           & \multicolumn{1}{c}{\textbf{Gaussian}}       & \multicolumn{1}{c}{\textbf{Impulse}}         & \multicolumn{1}{c}{\textbf{Bg.}}             & \multicolumn{1}{c}{\textbf{Occlusion}}       & \multicolumn{1}{c}{\textbf{LiDAR}}           & \multicolumn{1}{c}{\textbf{Inc.}}           & \multicolumn{1}{c}{\textbf{Dec.}}           & \multicolumn{1}{c}{\textbf{Cutout}}         & \multicolumn{1}{c}{\textbf{m\_OA}}\\ \hline
			\textbf{PointNet\cite{Pointnet}}                       &                               & {63.20}                    & {74.60}                    & {78.70}                    & {81.40}                    & {82.20}                    & {87.6}                    & {85.60}                    & {70.90}                    & {86.00}                       & {6.40}                      & {47.70}                     & {45.10}                     & {88.40}                    & {88.00}                      & {87.60}                    & {71.56}        \\
			\textbf{PointNet++\dag\cite{Pointnet++}}                &                               & {72.40}                    & {86.60}                    & {84.80}                    & {83.60}                    & {84.60}                    & {79.6}                    & {83.60}                    & {64.90}                    & {82.80}                     & {81.40}                     & {45.30}                     & {33.50}                     & {84.00}                      & {90.00}                      & {89.30}                    & {76.43}        \\
			\textbf{DGCNN\cite{DGCNN}}                          &                               & {80.90}                    & {87.90}                    & {86.90}                    & {85.50}                    & {86.00}                      & {85.4}                    & {83.40}                    & {75.10}                    & {80.90}                     & {46.90}                     & {40.80}                     & {19.00}                       & {85.90}                    & {82.70}                    & {84.60}                    & {74.13}        \\
			\textbf{RSCNN\cite{RS-CNN}}                          &                               & {70.80}                    & {83.00}                      & {81.90}                    & {80.80}                    & {81.40}                    & {75.4}                    & {81.70}                    & {53.80}                    & {79.90}                     & {81.70}                     & {48.20}                     & {31.60}                     & {83.20}                    & {86.80}                    & {86.20}                    & {73.76}        \\
			\textbf{PCT\cite{guo2021pct}}                            &                               & {81.90}           & {88.50}                    & {87.60}                    & {87.00}                      & {87.40}                    & {87.90}                    & {86.10}                    & {60.90}                    & {82.60}                     & {42.10}                     & {43.40}                     & {23.30}                     & {88.20}                    & {85.70}                    & {85.50}                    & {74.54}        \\
			\textbf{SimpleView\cite{SimpleView}}                     & \multirow{-6}{*}{}& {69.30}                    & {81.50}           & {83.00}             & {82.10}                    & {82.80}                    & {85.50}                    & {85.80}                    & {75.40}                    & {82.30}                     & {53.20}                     & {44.50}                     & {17.80}                     & {86.30}                    & {82.80}                    & {79.90}                    & {72.81}        \\ \hline
			& \textbf{ST}                 & \underline{80.09}                                         & 84.90                                         & 86.54                                         & 86.43                                     & 87.51                                         & 87.67                                         & 90.67                                     & 85.61                                         & 72.60                                         & 77.95                                     & 41.01                                         & 31.78                                         & 71.59                                     & 80.58                                         & 83.74                                         & 76.58                              \\
			& \textbf{cutmix\_k\cite{Pointcutmix}}          & \cellcolor[HTML]{FED5D2}79.30(-0.79)           & \cellcolor[HTML]{EBF1DE}\underline{86.51(+1.61)}          & \cellcolor[HTML]{EBF1DE}\underline{87.72(+1.18)}          & \cellcolor[HTML]{EBF1DE}88.37(+1.94)          & \cellcolor[HTML]{EBF1DE}\underline{88.73(+1.22)}          & \cellcolor[HTML]{EBF1DE}87.89(+0.22)          & \cellcolor[HTML]{EBF1DE}\underline{91.55(+0.88)}          & \cellcolor[HTML]{EBF1DE}85.94(+0.33)          & \cellcolor[HTML]{EBF1DE}87.39(+14.79)          & \cellcolor[HTML]{EBF1DE}82.39(+4.44)           & \cellcolor[HTML]{EBF1DE}48.3(+7.29)            & \cellcolor[HTML]{EBF1DE}45.71(+13.93)          & \cellcolor[HTML]{EBF1DE}88.46(+16.87)         & \cellcolor[HTML]{EBF1DE}\underline{90.08(+9.5)}           & \cellcolor[HTML]{EBF1DE}\textbf{90.75(+7.01)}& \underline{81.94}                              \\
			& \textbf{cutmix\_r\cite{Pointcutmix}}          & \cellcolor[HTML]{FED5D2}78.39(-1.70)           & \cellcolor[HTML]{FED5D2}84.53(-0.37)          & \cellcolor[HTML]{FED5D2}85.97(-0.57)          & \cellcolor[HTML]{EBF1DE}88.07(+1.64)          & \cellcolor[HTML]{EBF1DE}88.31(+0.80)           & \cellcolor[HTML]{EBF1DE}\underline{89.67(+2.0)}           & \cellcolor[HTML]{EBF1DE}91.03(+0.36)          & \cellcolor[HTML]{EBF1DE}\underline{89.29(+3.68)}          & \cellcolor[HTML]{EBF1DE}\underline{90.19(+17.59)}          & \cellcolor[HTML]{FED5D2}74.87(-3.08)           & \cellcolor[HTML]{EBF1DE}43.78(+2.77)           & \cellcolor[HTML]{EBF1DE}\underline{46.26(+14.48)}          & \cellcolor[HTML]{EBF1DE}88.88(+17.29)         & \cellcolor[HTML]{EBF1DE}90.05(+9.47)          & \cellcolor[HTML]{EBF1DE}\underline{90.25(+6.51)}          & 81.3                               \\
			& \textbf{mixup\cite{Pointmixup}}              & \cellcolor[HTML]{FED5D2}78.74(-1.35)          & \cellcolor[HTML]{EBF1DE}86.06(+1.16)          & \cellcolor[HTML]{EBF1DE}87.15(+0.61)          & \cellcolor[HTML]{EBF1DE}88.17(+1.74)          & \cellcolor[HTML]{EBF1DE}88.7(+1.19)           & \cellcolor[HTML]{EBF1DE}88.75(+1.08)          & \cellcolor[HTML]{EBF1DE}91.45(+0.78)          & \cellcolor[HTML]{EBF1DE}87.5(+1.89)           & \cellcolor[HTML]{EBF1DE}84.92(+12.32)          & \cellcolor[HTML]{FED5D2}61.39(-16.56)          & \cellcolor[HTML]{EBF1DE}43.15(+2.14)           & \cellcolor[HTML]{EBF1DE}32.07(+0.29)           & \cellcolor[HTML]{EBF1DE}84.03(+12.44)         & \cellcolor[HTML]{EBF1DE}85.49(+4.91)          & \cellcolor[HTML]{EBF1DE}87.8(+4.06)           & 78.36                              \\
			& \textbf{rsmix\cite{RSMix}}              & \cellcolor[HTML]{FED5D2}63.78(-16.31)         & \cellcolor[HTML]{FED5D2}80.29(-4.61)          & \cellcolor[HTML]{FED5D2}80.49(-6.05)          & \cellcolor[HTML]{FED5D2}78.79(-7.64)          & \cellcolor[HTML]{FED5D2}79.97(-7.54)          & \cellcolor[HTML]{FED5D2}76.60(-11.07)          & \cellcolor[HTML]{EBF1DE}91.43(+0.76)          & \cellcolor[HTML]{FED5D2}69.81(-15.80)         & \cellcolor[HTML]{EBF1DE}88.14(+15.54)          & \cellcolor[HTML]{EBF1DE}\underline{84.01(+6.06)}           & \cellcolor[HTML]{EBF1DE}\underline{51.61(+10.60)}           & \cellcolor[HTML]{EBF1DE}45.53(+13.75)          & \cellcolor[HTML]{EBF1DE}\underline{88.51(+16.92)}         & \cellcolor[HTML]{EBF1DE}89.83(+9.25)          & \cellcolor[HTML]{EBF1DE}90.06(+6.32)          & 77.26                              \\
			\multirow{-6}{*}{\textbf{PointNet++\ddag}\cite{Pointnet++}}   & \textbf{DesenAT-sD(ours)}   & \cellcolor[HTML]{EBF1DE}\textbf{84.98(+4.89)}& \cellcolor[HTML]{EBF1DE}\textbf{91.62(+6.72)}& \cellcolor[HTML]{EBF1DE}\textbf{89.13(+2.59)}& \cellcolor[HTML]{EBF1DE}\textbf{90.05(+3.62)}& \cellcolor[HTML]{EBF1DE}\textbf{90.45(+2.94)}& \cellcolor[HTML]{EBF1DE}\textbf{91.64(+3.97)}& \cellcolor[HTML]{EBF1DE}\textbf{91.92(+1.25)}& \cellcolor[HTML]{EBF1DE}\textbf{91.57(+5.96)}& \cellcolor[HTML]{EBF1DE}\textbf{90.92(+18.32)}& \cellcolor[HTML]{EBF1DE}\textbf{88.53(+10.58)}& \cellcolor[HTML]{EBF1DE}\textbf{52.36(+11.35)}& \cellcolor[HTML]{EBF1DE}\textbf{54.12(+22.34)}& \cellcolor[HTML]{EBF1DE}\textbf{89.2(+17.61)}& \cellcolor[HTML]{EBF1DE}\textbf{90.4(+9.82)}& \cellcolor[HTML]{EBF1DE}90.21(+6.47)          & \textbf{85.14}                   \\ \hline
			& \textbf{ST}                 & \underline{80.86}                                         & 86.08                                         & 87.06                                         & 87.13                                         & 87.53                                         & 88.26                                         & 91.09                                         & 86.73                                         & 66.36                                         & 80.58                                         & 48.87                                          & 36.29                                         & 87.91                                         & 82.49                                         & 85.00                                         & 78.82                              \\
			& \textbf{cutmix\_k\cite{Pointcutmix}}          & \cellcolor[HTML]{FED5D2}74.8(-6.06)           & \cellcolor[HTML]{FED5D2}84.93(-1.15)          & \cellcolor[HTML]{FED5D2}86.28(-0.78)          & \cellcolor[HTML]{FED5D2}86.18(-0.95)          & \cellcolor[HTML]{FED5D2}86.78(-0.75)          & \cellcolor[HTML]{FED5D2}80.29(-7.97)          & \cellcolor[HTML]{EBF1DE}92.44(+1.35)          & \cellcolor[HTML]{FED5D2}75.53(-11.20)          & \cellcolor[HTML]{EBF1DE}88.05(+21.69)          & \cellcolor[HTML]{EBF1DE}\underline{85.37(+4.79)}           & \cellcolor[HTML]{EBF1DE}\textbf{54.13(+5.26)}& \cellcolor[HTML]{EBF1DE}49.68(+13.39)          & \cellcolor[HTML]{EBF1DE}91.06(+3.15)          & \cellcolor[HTML]{EBF1DE}90.32(+7.83)          & \cellcolor[HTML]{EBF1DE}\textbf{91.24(+6.24)}& 81.14                              \\
			& \textbf{cutmix\_r\cite{Pointcutmix}}          & \cellcolor[HTML]{FED5D2}78.26(-2.60)           & \cellcolor[HTML]{FED5D2}85.78(-0.30)           & \cellcolor[HTML]{EBF1DE}87.13(+0.07)          & \cellcolor[HTML]{EBF1DE}87.81(+0.68)          & \cellcolor[HTML]{EBF1DE}88.61(+1.08)          & \cellcolor[HTML]{FED5D2}87.71(-0.55)          & \cellcolor[HTML]{EBF1DE}\textbf{92.55(+1.46)}& \cellcolor[HTML]{FED5D2}86.6(-0.13)           & \cellcolor[HTML]{EBF1DE}\textbf{91.32(+24.96)}& \cellcolor[HTML]{EBF1DE}83.53(+2.95)           & \cellcolor[HTML]{FED5D2}46.16(-2.71)           & \cellcolor[HTML]{EBF1DE}\underline{50.53(+14.24)}          & \cellcolor[HTML]{EBF1DE}\underline{91.45(+3.54)}          & \cellcolor[HTML]{EBF1DE}\textbf{90.75(+8.26)}& \cellcolor[HTML]{EBF1DE}91.01(+6.01)          & \underline{82.61}                              \\
			& \textbf{mixup\cite{Pointmixup}}              & \cellcolor[HTML]{FED5D2}80.11(-0.75)          & \cellcolor[HTML]{EBF1DE}\underline{87.48(+1.4)}           & \cellcolor[HTML]{EBF1DE}\underline{87.93(+0.87)}          & \cellcolor[HTML]{EBF1DE}\underline{88.49(+1.36)}          & \cellcolor[HTML]{EBF1DE}\underline{88.95(+1.42)}          & \cellcolor[HTML]{EBF1DE}\underline{88.65(+0.39)}          & \cellcolor[HTML]{EBF1DE}91.68(+0.59)          & \cellcolor[HTML]{EBF1DE}\underline{87.92(+1.19)}          & \cellcolor[HTML]{EBF1DE}87.81(+21.45)          & \cellcolor[HTML]{FED5D2}69.64(-10.94)          & \cellcolor[HTML]{FED5D2}46.87(-2.00)            & \cellcolor[HTML]{EBF1DE}38.7(+2.41)            & \cellcolor[HTML]{EBF1DE}89.20(+1.29)           & \cellcolor[HTML]{EBF1DE}87.15(+4.66)          & \cellcolor[HTML]{EBF1DE}89.01(+4.01)          & 80.64                              \\
			& \textbf{rsmix\cite{RSMix}}              & \cellcolor[HTML]{FED5D2}68.02(-12.84)         & \cellcolor[HTML]{FED5D2}83.83(-2.25)          & \cellcolor[HTML]{FED5D2}83.34(-3.72)          & \cellcolor[HTML]{FED5D2}81.72(-5.41)          & \cellcolor[HTML]{FED5D2}82.81(-4.72)          & \cellcolor[HTML]{FED5D2}74.91(-13.35)         & \cellcolor[HTML]{EBF1DE}\underline{92.35(+1.26)}          & \cellcolor[HTML]{FED5D2}69.08(-17.65)         & \cellcolor[HTML]{EBF1DE}90.62(+24.26)          & \cellcolor[HTML]{EBF1DE}82.55(+1.97)           & \cellcolor[HTML]{FED5D2}46.7(-2.17)            & \cellcolor[HTML]{EBF1DE}\textbf{49.78(+13.49)}& \cellcolor[HTML]{EBF1DE}\textbf{91.12(+3.21)}& \cellcolor[HTML]{EBF1DE}\underline{90.73(+8.24)}          & \cellcolor[HTML]{EBF1DE}\textbf{91.24(+6.24)}& 78.59                              \\
			\multirow{-6}{*}{\textbf{PointNetMeta-S}\cite{pointNetMetaBase2023cvpr}}& \textbf{DesenAT-sD(ours)}   & \cellcolor[HTML]{EBF1DE}\textbf{81.31(+0.45)}& \cellcolor[HTML]{EBF1DE}\textbf{91.6(+5.52)}& \cellcolor[HTML]{EBF1DE}\textbf{87.94(+0.88)}& \cellcolor[HTML]{EBF1DE}\textbf{88.97(+1.84)}& \cellcolor[HTML]{EBF1DE}\textbf{89.31(+1.78)}& \cellcolor[HTML]{EBF1DE}\textbf{91.73(+3.47)}& \cellcolor[HTML]{EBF1DE}91.89(+0.80)           & \cellcolor[HTML]{EBF1DE}\textbf{91.87(+5.14)}& \cellcolor[HTML]{EBF1DE}\underline{91.19(+24.83)}          & \cellcolor[HTML]{EBF1DE}\textbf{89.93(+9.35)}& \cellcolor[HTML]{EBF1DE}\underline{52.42(+3.55)}           & \cellcolor[HTML]{EBF1DE}45.4(+9.11)            & \cellcolor[HTML]{EBF1DE}90.06(+2.15)          & \cellcolor[HTML]{EBF1DE}89.17(+6.68)          & \cellcolor[HTML]{EBF1DE}90.22(+5.22)          & \textbf{84.20}                    \\ \hline
			& \textbf{ST}                 & 84.4                                          & \underline{89.84}                                         & 89.34                                         & 88.09                                         & 88.70                                          & 89.00                                         & 92.05                                         & 87.44                                         & 80.51                                         & 90.21                                         & 42.59                                          & 33.15                                         & 88.95                                         & 86.23                                         & 86.53                                         & 81.14                              \\
			& \textbf{cutmix\_k\cite{Pointcutmix}}          & \cellcolor[HTML]{FED5D2}81.82(-2.58)          & \cellcolor[HTML]{FED5D2}87.02(-2.82)          & \cellcolor[HTML]{FED5D2}86.77(-2.57)          & \cellcolor[HTML]{FED5D2}87.52(-0.57)          & \cellcolor[HTML]{FED5D2}87.96(-0.74)          & \cellcolor[HTML]{FED5D2}87.62(-1.38)          & \cellcolor[HTML]{FED5D2}90.67(-1.38)          & \cellcolor[HTML]{FED5D2}86.38(-1.06)          & \cellcolor[HTML]{EBF1DE}85.07(+4.56)           & \cellcolor[HTML]{FED5D2}84.91(-5.30)            & \cellcolor[HTML]{FED5D2}37.85(-4.74)           & \cellcolor[HTML]{EBF1DE}36.3(+3.15)            & \cellcolor[HTML]{FED5D2}87.24(-1.71)          & \cellcolor[HTML]{EBF1DE}86.34(+0.11)          & \cellcolor[HTML]{EBF1DE}87.25(+0.72)          & 80.05                              \\
			& \textbf{cutmix\_r\cite{Pointcutmix}}          & \cellcolor[HTML]{FED5D2}82.79(-1.61)          & \cellcolor[HTML]{FED5D2}87.37(-2.47)          & \cellcolor[HTML]{FED5D2}87.49(-1.85)          & \cellcolor[HTML]{EBF1DE}\underline{88.67(+0.58)}          & \cellcolor[HTML]{EBF1DE}\underline{89.1(+0.4)}            & \cellcolor[HTML]{EBF1DE}\underline{90.02(+1.02)}          & \cellcolor[HTML]{FED5D2}91.66(-0.39)          & \cellcolor[HTML]{EBF1DE}\underline{89.18(+1.74)}          & \cellcolor[HTML]{EBF1DE}\underline{88.04(+7.53)}           & \cellcolor[HTML]{FED5D2}87.59(-2.62)           & \cellcolor[HTML]{FED5D2}42.23(-0.36)           & \cellcolor[HTML]{EBF1DE}39.47(+6.32)           & \cellcolor[HTML]{EBF1DE}89.74(+0.79)          & \cellcolor[HTML]{EBF1DE}88.74(+2.51)          & \cellcolor[HTML]{EBF1DE}86.56(+0.03)          & 81.91                              \\
			& \textbf{mixup\cite{Pointmixup}}              & \cellcolor[HTML]{EBF1DE}\underline{84.96(+0.56)}          & \cellcolor[HTML]{FED5D2}88.46(-1.38)          & \cellcolor[HTML]{FED5D2}88.19(-1.15)          & \cellcolor[HTML]{EBF1DE}88.53(+0.44)          & \cellcolor[HTML]{EBF1DE}88.96(+0.26)          & \cellcolor[HTML]{FED5D2}88.9(-0.10)            & \cellcolor[HTML]{FED5D2}90.83(-1.22)          & \cellcolor[HTML]{EBF1DE}88.02(+0.58)          & \cellcolor[HTML]{EBF1DE}83.41(+2.90)            & \cellcolor[HTML]{FED5D2}80.15(-10.06)          & \cellcolor[HTML]{FED5D2}39.21(-3.38)           & \cellcolor[HTML]{EBF1DE}40.09(+6.94)           & \cellcolor[HTML]{EBF1DE}88.95(+0.00)            & \cellcolor[HTML]{EBF1DE}86.58(+0.35)          & \cellcolor[HTML]{EBF1DE}86.89(+0.36)          & 80.81                              \\
			& \textbf{rsmix\cite{RSMix}}              & \cellcolor[HTML]{FED5D2}83.62(-0.78)          & \cellcolor[HTML]{FED5D2}89.67(-0.17)          & \cellcolor[HTML]{EBF1DE}\underline{89.80(+0.46)}           & \cellcolor[HTML]{EBF1DE}89.66(+1.57)          & \cellcolor[HTML]{EBF1DE}89.87(+1.17)          & \cellcolor[HTML]{EBF1DE}89.12(+0.12)          & \cellcolor[HTML]{EBF1DE}\underline{92.20(+0.15)}           & \cellcolor[HTML]{EBF1DE}87.76(+0.32)          & \cellcolor[HTML]{EBF1DE}86.13(+5.62)           & \cellcolor[HTML]{EBF1DE}\underline{90.28(+0.07)}           & \cellcolor[HTML]{EBF1DE}\underline{45.72(+3.13)}           & \cellcolor[HTML]{EBF1DE}\underline{42.55(+9.40)}            & \cellcolor[HTML]{EBF1DE}\underline{91.4(+2.45)}           & \cellcolor[HTML]{EBF1DE}\textbf{91.03(+4.80)}& \cellcolor[HTML]{EBF1DE}\textbf{91.38(+4.85)}& \underline{83.35}                              \\
			\multirow{-6}{*}{\textbf{APES\_global\cite{APES2023cvpr}}} & \textbf{DesenAT-sD(ours)}   & \cellcolor[HTML]{EBF1DE}\textbf{87.68(+3.28)}& \cellcolor[HTML]{EBF1DE}\textbf{92.5(+2.66)}& \cellcolor[HTML]{EBF1DE}\textbf{90.89(+1.55)}& \cellcolor[HTML]{EBF1DE}\textbf{90.65(+2.56)}& \cellcolor[HTML]{EBF1DE}\textbf{90.97(+2.27)}& \cellcolor[HTML]{EBF1DE}\textbf{92.56(+3.56)}& \cellcolor[HTML]{EBF1DE}\textbf{92.75(+0.7)}& \cellcolor[HTML]{EBF1DE}\textbf{92.39(+4.95)}& \cellcolor[HTML]{EBF1DE}\textbf{91.18(+10.67)}& \cellcolor[HTML]{EBF1DE}\textbf{91.87(+1.66)}& \cellcolor[HTML]{EBF1DE}\textbf{46.75(+4.16)}& \cellcolor[HTML]{EBF1DE}\textbf{41.61(+8.46)}& \cellcolor[HTML]{EBF1DE}\textbf{91.08(+2.13)}& \cellcolor[HTML]{EBF1DE}\underline{89.54(+3.31)}          & \cellcolor[HTML]{EBF1DE}\underline{89.45(+2.92)}          & \textbf{84.79}                   \\ \hline
			& \textbf{ST}                 & 84.94                                         & 89.32                                         & 89.50                                         & 88.50                                          & 89.09                                         & 89.27                                         & 91.80                                          & 88.04                                         & 82.70                                           & \underline{90.74}                                          & \underline{44.10}                                           & 37.16                                          & 89.12                                         & 86.41                                         & 87.01                                         & 81.85                              \\
			& \textbf{cutmix\_k\cite{Pointcutmix}}          & \cellcolor[HTML]{FED5D2}80.41(-4.53)          & \cellcolor[HTML]{FED5D2}84.73(-4.59)          & \cellcolor[HTML]{FED5D2}84.37(-5.13)          & \cellcolor[HTML]{FED5D2}85.31(-3.19)          & \cellcolor[HTML]{FED5D2}86.08(-3.01)          & \cellcolor[HTML]{FED5D2}84.98(-4.29)          & \cellcolor[HTML]{FED5D2}88.38(-3.42)          & \cellcolor[HTML]{FED5D2}83.77(-4.27)          & \cellcolor[HTML]{FED5D2}80.73(-1.97)           & \cellcolor[HTML]{FED5D2}76.49(-14.25)          & \cellcolor[HTML]{FED5D2}41.17(-2.93)           & \cellcolor[HTML]{EBF1DE}40.69(+3.53)           & \cellcolor[HTML]{FED5D2}86.03(-3.09)          & \cellcolor[HTML]{FED5D2}84.32(-2.09)          & \cellcolor[HTML]{FED5D2}85.75(-1.26)          & 78.21                              \\
			& \textbf{cutmix\_r\cite{Pointcutmix}}          & \cellcolor[HTML]{FED5D2}80.88(-4.06)          & \cellcolor[HTML]{FED5D2}84.17(-5.15)          & \cellcolor[HTML]{FED5D2}85.59(-3.91)          & \cellcolor[HTML]{FED5D2}86.35(-2.15)          & \cellcolor[HTML]{FED5D2}86.65(-2.44)          & \cellcolor[HTML]{FED5D2}87.12(-2.15)          & \cellcolor[HTML]{FED5D2}88.66(-3.14)          & \cellcolor[HTML]{FED5D2}85.97(-2.07)          & \cellcolor[HTML]{FED5D2}82.4(-0.30)             & \cellcolor[HTML]{FED5D2}78.34(-12.40)           & \cellcolor[HTML]{FED5D2}37.37(-6.73)           & \cellcolor[HTML]{EBF1DE}39.96(+2.80)            & \cellcolor[HTML]{FED5D2}87.43(-1.69)          & \cellcolor[HTML]{FED5D2}85.78(-0.63)          & \cellcolor[HTML]{FED5D2}85.76(-1.25)          & 78.83                              \\
			& \textbf{mixup\cite{Pointmixup}}              & \cellcolor[HTML]{EBF1DE}\textbf{86.94(+2.00)}& \cellcolor[HTML]{EBF1DE}\underline{90.26(+0.94)}          & \cellcolor[HTML]{EBF1DE}\underline{89.83(+0.33)}          & \cellcolor[HTML]{EBF1DE}\underline{90.02(+1.52)}          & \cellcolor[HTML]{EBF1DE}\underline{90.50(+1.41)}           & \cellcolor[HTML]{EBF1DE}\underline{90.67(+1.40)}           & \cellcolor[HTML]{EBF1DE}91.84(+0.04)          & \cellcolor[HTML]{EBF1DE}\underline{90.39(+2.35)}          & \cellcolor[HTML]{EBF1DE}87.0(+4.30)             & \cellcolor[HTML]{FED5D2}83.81(-6.93)           & \cellcolor[HTML]{FED5D2}43.44(-0.66)           & \cellcolor[HTML]{EBF1DE}41.75(+4.59)           & \cellcolor[HTML]{EBF1DE}89.97(+0.85)          & \cellcolor[HTML]{EBF1DE}88.01(+1.60)           & \cellcolor[HTML]{EBF1DE}\underline{89.35(+2.34)}          & 82.92                              \\
			& \textbf{rsmix\cite{RSMix}}              & \cellcolor[HTML]{FED5D2}83.12(-1.82)          & \cellcolor[HTML]{EBF1DE}89.81(+0.49)          & \cellcolor[HTML]{EBF1DE}89.79(+0.29)          & \cellcolor[HTML]{EBF1DE}89.89(+1.39)          & \cellcolor[HTML]{EBF1DE}90.33(+1.24)          & \cellcolor[HTML]{EBF1DE}89.37(+0.10)           & \cellcolor[HTML]{EBF1DE}\underline{92.14(+0.34)}          & \cellcolor[HTML]{EBF1DE}88.36(+0.32)          & \cellcolor[HTML]{EBF1DE}\underline{88.22(+5.52)}           & \cellcolor[HTML]{FED5D2}90.17(-0.57)           & \cellcolor[HTML]{FED5D2}43.54(-0.56)           & \cellcolor[HTML]{EBF1DE}\textbf{43.06(+5.90)} & \cellcolor[HTML]{EBF1DE}\textbf{91.4(+2.28)}& \cellcolor[HTML]{EBF1DE}\textbf{90.93(+4.52)}& \cellcolor[HTML]{EBF1DE}\textbf{91.29(+4.28)}& \underline{83.43}                              \\
			\multirow{-6}{*}{\textbf{APES\_local\cite{APES2023cvpr}}}  & \textbf{DesenAT-sD(ours)}   & \cellcolor[HTML]{EBF1DE}\underline{85.88(+0.94)}          & \cellcolor[HTML]{EBF1DE}\textbf{92.16(+2.84)}& \cellcolor[HTML]{EBF1DE}\textbf{90.33(+0.83)}& \cellcolor[HTML]{EBF1DE}\textbf{90.56(+2.06)}& \cellcolor[HTML]{EBF1DE}\textbf{91.06(+1.97)}& \cellcolor[HTML]{EBF1DE}\textbf{91.98(+2.71)}& \cellcolor[HTML]{EBF1DE}\textbf{92.41(+0.61)}& \cellcolor[HTML]{EBF1DE}\textbf{91.71(+3.67)}& \cellcolor[HTML]{EBF1DE}\textbf{89.78(+7.08)}& \cellcolor[HTML]{EBF1DE}\textbf{91.39(+0.65)}& \cellcolor[HTML]{EBF1DE}\textbf{45.4(+1.30)}  & \cellcolor[HTML]{EBF1DE}\underline{41.82(+4.66)}           & \cellcolor[HTML]{EBF1DE}\underline{90.37(+1.25)}          & \cellcolor[HTML]{EBF1DE}\underline{88.87(+2.46)}          & \cellcolor[HTML]{EBF1DE}89.04(+2.03)          & \textbf{84.18}\\ \hline
		\end{tabular}
	}
\end{table}

% 7 OA
\begin{table}[!h]
	\centering
	\caption{The mOA(\%) of different model architectures on PointCloud-C with different training strategies. Bold: best in column. Underline: second best in column. \ddag: multi-scale inference from \cite{RepSurf}, ST: standard training. Red: Above ST, Green: Below ST.}\label{tab 2}
	\resizebox{\linewidth}{!}{
		\begin{tabular}{llrrrrrrrr} \hline
			&                               & \multicolumn{1}{c}{\textbf{add\_global}}    & \multicolumn{1}{c}{\textbf{add\_local}}     & \multicolumn{1}{c}{\textbf{dropout\_global}} & \multicolumn{1}{c}{\textbf{dropout\_local}}  & \multicolumn{1}{c}{\textbf{jitter}}          & \multicolumn{1}{c}{\textbf{rotate}}         & \multicolumn{1}{c}{\textbf{scale}}          & \multicolumn{1}{c}{\textbf{m\_OA}}\\ \hline
			\textbf{DGCNN\cite{DGCNN}}                          &                               & {70.50}                    & {72.50}                    & {75.20}                     & {79.30}                     & {68.40}                     & {78.50}                    & {90.60}                    & {76.43}      \\
			\textbf{PointNet\cite{Pointnet}}                       &                               & {81.90}                    & {72.70}                    & {84.10}                     & {62.70}                     & {62.80}                     & {69.80}                    & {91.80}                    & {75.11}      \\
			\textbf{RSCNN\cite{RS-CNN}}                          &                               & {79.00}                      & {68.30}                    & {80.00}                       & {68.60}                     & {63.00}                       & {68.20}                    & {89.90}                    & {73.86}      \\
			\textbf{SimpleView\cite{SimpleView}}                     &                               & {71.00}                      & {76.80}                    & {69.20}                     & {71.90}                     & {77.40}                     & {71.70}                    & {91.80}                    & {75.69}      \\
			\textbf{GDANet\cite{xu2021learning}}                         &                               & {74.30}                    & {71.50}                    & {80.30}                     & {81.50}                     & {73.50}                     & {78.90}                    & {92.20}                    & {78.89}      \\
			\textbf{CurveNet\cite{CurveNet}}                       &                               & {60.30}                    & {72.50}                    & {82.40}                     & {78.80}                     & {77.10}                     & {82.60}                    & {91.80}                    & {77.93}      \\
			\textbf{PAConv\cite{PAConv}}                         &                               & {68.00}                      & {64.30}                    & {75.20}                     & {79.20}                     & {53.70}                     & {79.20}                    & {91.50}                    & {73.01}      \\
			\textbf{PCT\cite{guo2021pct}}                            &                               & {77.00}                      & {61.90}                    & {86.90}                     & {79.30}                     & {72.50}                     & {77.60}                    & {91.80}                    & {78.14}      \\
			\textbf{RPC\cite{Pointcloud-C}}                            & \multirow{-9}{*}{}& {72.60}                    & {72.20}                    & {87.80}                     & {83.50}                     & {71.80}                     & {76.80}                    & {92.10}                    & {79.54}      \\ \hline
			& \textbf{ST}                 & 91.37                                         & 90.58                                         & 71.54                                          & 54.92                                          & 75.84                                          & \underline{75.16}                                         & 81.03                                         & 77.21                            \\
			& \textbf{cutmix\_k\cite{Pointcutmix}}          & \cellcolor[HTML]{EBF1DE}92.05(+0.68)          & \cellcolor[HTML]{EBF1DE}91.56(+0.98)          & \cellcolor[HTML]{EBF1DE}83.15(+11.61)          & \cellcolor[HTML]{EBF1DE}\underline{79.37(+24.45)}          & \cellcolor[HTML]{FED8D2}75.38(-0.46)           & \cellcolor[HTML]{FED8D2}72.92(-2.24)          & \cellcolor[HTML]{EBF1DE}81.93(+0.90)           & 82.34                            \\
			& \textbf{cutmix\_r\cite{Pointcutmix}}          & \cellcolor[HTML]{EBF1DE}91.71(+0.34)          & \cellcolor[HTML]{EBF1DE}91.13(+0.55)          & \cellcolor[HTML]{EBF1DE}\underline{89.39(+17.85)}          & \cellcolor[HTML]{EBF1DE}71.64(+16.72)          & \cellcolor[HTML]{EBF1DE}\underline{85.54(+9.7)}            & \cellcolor[HTML]{FED8D2}71.95(-3.21)          & \cellcolor[HTML]{EBF1DE}81.36(+0.33)          & \underline{83.25}                            \\
			& \textbf{mixup\cite{Pointmixup}}              & \cellcolor[HTML]{EBF1DE}\underline{92.29(+0.92)}          & \cellcolor[HTML]{EBF1DE}91.76(+1.18)          & \cellcolor[HTML]{FED8D2}70.92(-0.62)           & \cellcolor[HTML]{EBF1DE}65.42(+10.5)           & \cellcolor[HTML]{EBF1DE}77.16(+1.32)           & \cellcolor[HTML]{FED8D2}73.92(-1.24)          & \cellcolor[HTML]{EBF1DE}84.22(+3.19)          & 79.38                            \\
			& \textbf{rsmix\cite{RSMix}}              & \cellcolor[HTML]{EBF1DE}\textbf{92.32(+0.95)}& \cellcolor[HTML]{EBF1DE}\textbf{92.19(+1.61)}& \cellcolor[HTML]{EBF1DE}79.04(+7.50)            & \cellcolor[HTML]{EBF1DE}\textbf{79.52(+24.6)}& \cellcolor[HTML]{FED8D2}52.33(-23.51)          & \cellcolor[HTML]{FED8D2}61.46(-13.7)          & \cellcolor[HTML]{EBF1DE}\underline{84.90(+3.87)}           & 77.39                            \\
			\multirow{-6}{*}{\textbf{PointNet++\ddag}\cite{Pointnet++}}   & \textbf{DesenAT-sD(ours)}   & \cellcolor[HTML]{EBF1DE}92.07(+0.7)           & \cellcolor[HTML]{EBF1DE}\underline{91.91(+1.33)}          & \cellcolor[HTML]{EBF1DE}\textbf{89.53(+17.99)}& \cellcolor[HTML]{EBF1DE}71.06(+16.14)          & \cellcolor[HTML]{EBF1DE}\textbf{90.69(+14.85)}& \cellcolor[HTML]{EBF1DE}\textbf{78.6(+3.44)}& \cellcolor[HTML]{EBF1DE}\textbf{88.31(+7.28)}& \textbf{86.02}                 \\ \hline
			& \textbf{ST}                 & 91.29                                         & 91.14                                         & 65.62                                          & 73.35                                          & 73.88                                          & \underline{75.01}                                         & 80.62                                         & 78.7                             \\
			& \textbf{cutmix\_k\cite{Pointcutmix}}          & \cellcolor[HTML]{EBF1DE}\textbf{93.11(+1.82)}& \cellcolor[HTML]{EBF1DE}\textbf{92.56(+1.42)}& \cellcolor[HTML]{EBF1DE}76.11(+10.49)          & \cellcolor[HTML]{EBF1DE}\underline{87.49(+14.14)}          & \cellcolor[HTML]{FED8D2}56.93(-16.95)          & \cellcolor[HTML]{FED8D2}70.32(-4.69)          & \cellcolor[HTML]{EBF1DE}83.61(+2.99)          & 80.02                            \\
			& \textbf{cutmix\_r\cite{Pointcutmix}}          & \cellcolor[HTML]{EBF1DE}92.39(+1.10)           & \cellcolor[HTML]{EBF1DE}92.33(+1.19)          & \cellcolor[HTML]{EBF1DE}\underline{85.32(+19.70)}           & \cellcolor[HTML]{EBF1DE}82.74(+9.39)           & \cellcolor[HTML]{EBF1DE}78.24(+4.36)           & \cellcolor[HTML]{FED8D2}72.18(-2.83)          & \cellcolor[HTML]{EBF1DE}82.59(+1.97)          & \underline{83.68}                            \\
			& \textbf{mixup\cite{Pointmixup}}              & \cellcolor[HTML]{EBF1DE}92.55(+1.26)          & \cellcolor[HTML]{EBF1DE}92.02(+0.88)          & \cellcolor[HTML]{EBF1DE}73.47(+7.85)           & \cellcolor[HTML]{EBF1DE}80.52(+7.17)           & \cellcolor[HTML]{EBF1DE}\underline{80.22(+6.34)}           & \cellcolor[HTML]{FED8D2}74.34(-0.67)          & \cellcolor[HTML]{EBF1DE}\underline{84.94(+4.32)}          & 82.58                            \\
			& \textbf{rsmix\cite{RSMix}}              & \cellcolor[HTML]{EBF1DE}\underline{92.74(+1.45)}          & \cellcolor[HTML]{EBF1DE}\underline{92.38(+1.24)}          & \cellcolor[HTML]{EBF1DE}81.49(+15.87)          & \cellcolor[HTML]{EBF1DE}\textbf{89.12(+15.77)}& \cellcolor[HTML]{FED8D2}50.99(-22.89)          & \cellcolor[HTML]{FED8D2}65.24(-9.77)          & \cellcolor[HTML]{EBF1DE}84.72(+4.10)           & 79.53                            \\
			\multirow{-6}{*}{\textbf{PointNetMeta-S}\cite{pointNetMetaBase2023cvpr}}& \textbf{DesenAT-sD(ours)}   & \cellcolor[HTML]{EBF1DE}92.58(+1.29)          & \cellcolor[HTML]{EBF1DE}92.11(+0.97)          & \cellcolor[HTML]{EBF1DE}\textbf{86.43(+20.81)}& \cellcolor[HTML]{EBF1DE}78.46(+5.11)           & \cellcolor[HTML]{EBF1DE}\textbf{91.34(+17.46)}& \cellcolor[HTML]{EBF1DE}\textbf{76.34(+1.33)}& \cellcolor[HTML]{EBF1DE}\textbf{90.04(+9.42)}& \textbf{86.76}                 \\ \hline
			& \textbf{ST}                 & \underline{93.23}                                         & \underline{92.88}                                         & 79.91                                          & 79.40                                           & 78.98                                          & 80.08                                         & \underline{84.62}                                         & 84.16                            \\
			& \textbf{cutmix\_k\cite{Pointcutmix}}          & \cellcolor[HTML]{FED5D2}91.75(-1.48)          & \cellcolor[HTML]{FED5D2}91.31(-1.57)          & \cellcolor[HTML]{FED5D2}73.55(-6.36)           & \cellcolor[HTML]{EBF1DE}\underline{81.95(+2.55)}           & \cellcolor[HTML]{FED8D2}78.2(-0.78)            & \cellcolor[HTML]{FED8D2}77.36(-2.72)          & \cellcolor[HTML]{FED8D2}81.52(-3.10)           & 82.23                            \\
			& \textbf{cutmix\_r\cite{Pointcutmix}}          & \cellcolor[HTML]{FED8D2}92.58(-0.65)          & \cellcolor[HTML]{FED8D2}91.89(-0.99)          & \cellcolor[HTML]{EBF1DE}\underline{88.96(+9.05)}           & \cellcolor[HTML]{FED8D2}77.51(-1.89)           & \cellcolor[HTML]{EBF1DE}\underline{85.93(+6.95)}           & \cellcolor[HTML]{FED8D2}78.37(-1.71)          & \cellcolor[HTML]{FED8D2}\underline{82.62(-2.00)}           & 85.41                            \\
			& \textbf{mixup\cite{Pointmixup}}              & \cellcolor[HTML]{FED5D2}91.37(-1.86)          & \cellcolor[HTML]{FED8D2}90.85(-2.03)          & \cellcolor[HTML]{FED8D2}75.17(-4.74)           & \cellcolor[HTML]{FED5D2}75.99(-3.41)           & \cellcolor[HTML]{EBF1DE}82.89(+3.91)           & \cellcolor[HTML]{EBF1DE}\underline{81.09(+1.01)}          & \cellcolor[HTML]{FED8D2}82.07(-2.55)          & 82.78                            \\
			& \textbf{rsmix\cite{RSMix}}              & \cellcolor[HTML]{FED8D2}92.79(-0.44)          & \cellcolor[HTML]{FED8D2}92.47(-0.41)          & \cellcolor[HTML]{EBF1DE}\textbf{89.78(+9.87)}& \cellcolor[HTML]{EBF1DE}\textbf{88.44(+9.04)}& \cellcolor[HTML]{FED8D2}78.0(-0.98)            & \cellcolor[HTML]{FED8D2}78.27(-1.81)          & \cellcolor[HTML]{FED8D2}84.17(-0.45)          & \underline{86.27}                            \\
			\multirow{-6}{*}{\textbf{APES\_global}\cite{APES2023cvpr}} & \textbf{DesenAT-sD(ours)}   & \cellcolor[HTML]{EBF1DE}\textbf{93.27(+0.04)}& \cellcolor[HTML]{EBF1DE}\textbf{93.07(+0.19)}& \cellcolor[HTML]{EBF1DE}87.85(+7.94)           & \cellcolor[HTML]{EBF1DE}81.04(+1.64)           & \cellcolor[HTML]{EBF1DE}\textbf{91.61(+12.63)}& \cellcolor[HTML]{EBF1DE}\textbf{82.87(+2.79)}& \cellcolor[HTML]{EBF1DE}\textbf{90.15(+5.53)}& \textbf{88.55}                 \\ \hline
			& \textbf{ST}                 & 92.62                                         & 92.31                                         & 80.30                                           & 79.48                                          & 80.76                                          & 81.07                                         & \underline{85.23}                                         & 84.54                            \\
			& \textbf{cutmix\_k\cite{Pointcutmix}}          & \cellcolor[HTML]{FED8D2}89.43(-3.19)          & \cellcolor[HTML]{FED8D2}88.64(-3.67)          & \cellcolor[HTML]{FED8D2}73.01(-7.29)           & \cellcolor[HTML]{FED8D2}78.15(-1.33)           & \cellcolor[HTML]{FED8D2}74.96(-5.80)            & \cellcolor[HTML]{FED8D2}75.47(-5.6)           & \cellcolor[HTML]{FED8D2}78.14(-7.09)          & 79.69                            \\
			& \textbf{cutmix\_r\cite{Pointcutmix}}          & \cellcolor[HTML]{FED8D2}89.72(-2.9)           & \cellcolor[HTML]{FED8D2}88.59(-3.72)          & \cellcolor[HTML]{EBF1DE}81.85(+1.55)           & \cellcolor[HTML]{FED5D2}74.23(-5.25)           & \cellcolor[HTML]{EBF1DE}81.25(+0.49)           & \cellcolor[HTML]{FED8D2}76.54(-4.53)          & \cellcolor[HTML]{FED8D2}78.7(-6.53)           & 81.55                            \\
			& \textbf{mixup\cite{Pointmixup}}              & \cellcolor[HTML]{FED8D2}92.56(-0.06)          & \cellcolor[HTML]{FED8D2}92.13(-0.18)          & \cellcolor[HTML]{EBF1DE}81.55(+1.25)           & \cellcolor[HTML]{EBF1DE}\underline{81.39(+1.91)}           & \cellcolor[HTML]{EBF1DE}\underline{87.64(+6.88)}           & \cellcolor[HTML]{EBF1DE}\textbf{83.23(+2.16)}& \cellcolor[HTML]{FED8D2}84.37(-0.86)          & 86.12                            \\
			& \textbf{rsmix\cite{RSMix}}              & \cellcolor[HTML]{EBF1DE}\underline{92.78(+0.16)}          & \cellcolor[HTML]{EBF1DE}\underline{92.57(+0.26)}          & \cellcolor[HTML]{EBF1DE}\textbf{89.13(+8.83)}& \cellcolor[HTML]{EBF1DE}\textbf{88.01(+8.53)}& \cellcolor[HTML]{FED8D2}80.49(-0.27)           & \cellcolor[HTML]{FED8D2}77.79(-3.28)          & \cellcolor[HTML]{FED8D2}84.21(-1.02)          & \underline{86.43}                            \\
			\multirow{-6}{*}{\textbf{APES\_local}\cite{APES2023cvpr}}  & \textbf{DesenAT-sD(ours)}   & \cellcolor[HTML]{EBF1DE}\textbf{93.14(+0.52)}& \cellcolor[HTML]{EBF1DE}\textbf{92.9(+0.59)}& \cellcolor[HTML]{EBF1DE}\underline{85.68(+5.38)}           & \cellcolor[HTML]{EBF1DE}80.95(+1.47)           & \cellcolor[HTML]{EBF1DE}\textbf{90.25(+9.49)}& \cellcolor[HTML]{EBF1DE}\underline{81.41(+0.34)}          & \cellcolor[HTML]{EBF1DE}\textbf{88.75(+3.52)}& \textbf{87.58}\\ \hline
		\end{tabular}
	}
\end{table}

To evaluate the effectiveness and superiority of our proposed method (DesenAT-sD), we conducted comparisons with four mainstream models (PointNet++, PointNetMeta-S, APES\_global and APES\_Local) and four popular training methods (Cutmix-k, Cutout, Mixup, and Rsmix). PointNet++ (msg) and PointNetMeta-S are models based on the MLP framework, and APES\_global and APES\_Local are models based on the transformer framework. The results on ModelNet40-C and PointCloud-C datasets are presented in Tabs 1-4. Owing to space limitations, we placed the results of the evaluations for various levels of corruption in the Appendix. (Figs. \ref{Fig7}, \ref{Fig8}, \ref{Fig9}, \ref{Fig10}, \ref{Fig11}, \ref{Fig12}, \ref{Fig13}, and \ref{Fig14}).

In Tabs. \ref{tab 1} and \ref{tab 2}, we provide the performance differences between the mainstream and standard training methods under different benchmarks (within parentheses). The red regions indicate a decrease in accuracy compared to ST after applying the respective training methods, whereas the green regions indicate an improvement in accuracy relative to standard training methods. Notably, although these training methods (Cutmix-k, Cutout, Mixup, Rsmix) exhibit high performance on specific benchmarks, on some benchmarks, their performance is inferior to that of the ST method (for example, the Rsmix algorithm achieves the highest performance on Dropout\_global, however, it experiences significant loss compared to ST on the Jitter benchmark). Conversely, the DesenAT-sD method consistently outperforms the ST method on all benchmarks. Additionally, in terms of the average accuracy, DesenAT-sD outperforms other training methods on the four mainstream models. This indicates that the proposed method demonstrates a degree of universality and competitiveness on corrupted point clouds.

% 15 CE
\begin{table}[!th]
	\centering
	\caption{CE of different model architectures on ModelNet40-C with different training strategies. Bold: best in column. Underline: second best in column. \ddag: multi-scale inference from \cite{RepSurf}, ST: standard training.}\label{tab 3}
	\resizebox{\linewidth}{!}{%
		\begin{tabular}{llrrrrrrrrrrrrrrrr} 
			\hline
			\multicolumn{1}{l}{}                  & \multicolumn{1}{l}{}& \textbf{Rotation}& \textbf{Shear}  & \textbf{FFD}    & \textbf{RBF}    & \textbf{Lnv.RBF}& \textbf{Uniform}& \textbf{Ups.}   & \textbf{Gaussian}& \textbf{Impulse}& \textbf{Bg.}    & \textbf{Occlusion}& \textbf{LiDAR}  & \textbf{Inc.}   & \textbf{Dec.}   & \textbf{Cutout} & \textbf{m\_CE}  \\ \hline
			\multirow{6}{*}{\textbf{PointNet++\ddag}\cite{Pointnet++}}& \textbf{ST}        & 1.0(2)            & 1.0(4)            & 1.0(4)            & 1.0(5)            & 1.0(5)            & 1.0(5)            & 1.0(6)            & 1.0(5)            & 1.0(6)            & 1.0(4)            & 1.0(6)             & 1.0(6)            & 1.0(6)            & 1.0(6)            & 1.0(6)            & 1.0(5)            \\
			& \textbf{cutmix\_k\cite{Pointcutmix}}& 1.04(3)           & 0.894(2)          & 0.913(2)          & 0.857(2)          & 0.902(2)          & 0.982(4)          & 0.906(2)          & 0.977(4)          & 0.46(4)           & 0.799(3)          & 0.876(3)           & 0.796(3)          & 0.406(4)          & 0.511(2)          & \textbf{0.569(1)}& 0.793(2)          \\
			& \textbf{cutmix\_r\cite{Pointcutmix}} & 1.085(5)          & 1.025(5)          & 1.042(5)          & 0.879(4)          & 0.936(4)          & 0.838(2)          & 0.962(5)          & 0.744(2)          & 0.358(2)          & 1.14(5)           & 0.953(4)           & 0.788(2)          & 0.391(2)          & 0.512(3)          & 0.6(2)            & 0.817(3)          \\
			& \textbf{mixup\cite{Pointmixup}}     & 1.068(4)          & 0.923(3)          & 0.955(3)          & 0.872(3)          & 0.905(3)          & 0.912(3)          & 0.917(3)          & 0.868(3)          & 0.55(5)           & 1.751(6)          & 0.964(5)           & 0.996(5)          & 0.562(5)          & 0.747(5)          & 0.751(5)          & 0.916(4)          \\
			& \textbf{rsmix\cite{RSMix}}      & 1.819(6)          & 1.305(6)          & 1.45(6)           & 1.562(6)          & 1.604(6)          & 1.897(6)          & 0.919(4)          & 2.098(6)          & 0.433(3)          & 0.725(2)          & 0.82(2)            & 0.799(4)          & 0.404(3)          & 0.524(4)          & 0.612(4)          & 1.131(6)          \\
			& \textbf{DesenAT-sD(ours)}& \textbf{0.755(1)}& \textbf{0.555(1)}& \textbf{0.808(1)}& \textbf{0.733(1)}& \textbf{0.765(1)}& \textbf{0.677(1)}& \textbf{0.866(1)}& \textbf{0.586(1)}& \textbf{0.331(1)}& \textbf{0.52(1)}& \textbf{0.808(1)}& \textbf{0.673(1)}& \textbf{0.38(1)}& \textbf{0.494(1)}& 0.602(3)          & \textbf{0.637(1)}\\ \hline
			\multirow{6}{*}{\textbf{PointNetMeta-S}\cite{pointNetMetaBase2023cvpr}}& \textbf{ST}        & 0.961(2)          & 0.922(3)          & 0.961(4)          & 0.948(4)          & 0.999(4)          & 0.952(3)          & 0.955(6)          & 0.922(3)          & 1.228(6)          & 0.881(5)          & 0.867(3)           & 0.934(6)          & 0.426(6)          & 0.902(6)          & 0.923(6)          & 0.919(5)          \\
			& \textbf{cutmix\_k\cite{Pointcutmix}}& 1.266(5)          & 0.998(5)          & 1.019(5)          & 1.018(5)          & 1.058(5)          & 1.598(5)          & 0.811(2)          & 1.7(5)            & 0.436(4)          & 0.664(2)          & \textbf{0.778(1)}& 0.738(3)          & 0.315(3)          & 0.498(3)          & \textbf{0.539(1)}& 0.896(4)          \\
			& \textbf{cutmix\_r\cite{Pointcutmix}} & 1.092(4)          & 0.942(4)          & 0.956(3)          & 0.898(3)          & 0.912(3)          & 0.997(4)          & \textbf{0.799(1)}& 0.931(4)          & \textbf{0.317(1)}& 0.747(3)          & 0.913(6)           & \textbf{0.725(1)}& \textbf{0.301(1)}& \textbf{0.476(1)}& 0.553(3)          & 0.771(2)          \\
			& \textbf{mixup\cite{Pointmixup}}     & 0.999(3)          & 0.829(2)          & 0.897(2)          & 0.848(2)          & 0.885(2)          & 0.92(2)           & 0.892(5)          & 0.84(2)           & 0.445(5)          & 1.377(6)          & 0.901(4)           & 0.899(5)          & 0.38(5)           & 0.662(5)          & 0.676(5)          & 0.83(3)           \\
			& \textbf{rsmix\cite{RSMix}}      & 1.606(6)          & 1.071(6)          & 1.238(6)          & 1.347(6)          & 1.376(6)          & 2.034(6)          & 0.82(3)           & 2.148(6)          & 0.342(3)          & 0.791(4)          & 0.903(5)           & 0.736(2)          & 0.313(2)          & 0.477(2)          & 0.539(2)          & 1.049(6)          \\
			& \textbf{DesenAT-sD(ours)}& \textbf{0.939(1)}& \textbf{0.556(1)}& \textbf{0.896(1)}& \textbf{0.813(1)}& \textbf{0.856(1)}& \textbf{0.671(1)}& 0.87(4)           & \textbf{0.565(1)}& 0.322(2)          & \textbf{0.457(1)}& 0.807(2)           & 0.8(4)            & 0.35(4)           & 0.558(4)          & 0.602(4)          & \textbf{0.671(1)}\\ \hline
			\multirow{6}{*}{\textbf{APES\_global}\cite{APES2023cvpr}} & \textbf{ST}        & 0.784(3)          & 0.673(2)          & 0.792(3)          & 0.878(5)          & 0.905(5)          & 0.892(4)          & 0.852(3)          & 0.873(5)          & 0.711(6)          & 0.444(3)          & 0.973(3)           & 0.98(6)           & 0.389(4)          & 0.709(6)          & 0.828(6)          & 0.779(4)          \\
			& \textbf{cutmix\_k\cite{Pointcutmix}}& 0.913(6)          & 0.86(6)           & 0.983(6)          & 0.92(6)           & 0.964(6)          & 1.004(6)          & 1.0(6)            & 0.946(6)          & 0.545(4)          & 0.684(5)          & 1.054(6)           & 0.934(5)          & 0.449(6)          & 0.703(5)          & 0.785(3)          & 0.85(6)           \\
			& \textbf{cutmix\_r\cite{Pointcutmix}} & 0.865(5)          & 0.837(5)          & 0.93(5)           & 0.835(3)          & 0.873(3)          & 0.809(2)          & 0.894(4)          & 0.752(2)          & 0.436(2)          & 0.563(4)          & 0.979(4)           & 0.887(4)          & 0.361(3)          & 0.579(3)          & 0.827(5)          & 0.762(3)          \\
			& \textbf{mixup\cite{Pointmixup}}     & 0.755(2)          & 0.764(4)          & 0.878(4)          & 0.845(4)          & 0.884(4)          & 0.9(5)            & 0.983(5)          & 0.832(3)          & 0.605(5)          & 0.9(6)            & 1.031(5)           & 0.878(3)          & 0.389(5)          & 0.691(4)          & 0.807(4)          & 0.809(5)          \\
			& \textbf{rsmix\cite{RSMix}}      & 0.823(4)          & 0.684(3)          & 0.758(2)          & 0.762(2)          & 0.811(2)          & 0.882(3)          & 0.837(2)          & 0.85(4)           & 0.506(3)          & 0.441(2)          & 0.92(2)            & \textbf{0.842(1)}& \textbf{0.303(1)}& \textbf{0.462(1)}& \textbf{0.53(1)}& 0.694(2)          \\
			& \textbf{DesenAT-sD(ours)}& \textbf{0.619(1)}& \textbf{0.497(1)}& \textbf{0.677(1)}& \textbf{0.689(1)}& \textbf{0.723(1)}& \textbf{0.603(1)}& \textbf{0.778(1)}& \textbf{0.529(1)}& \textbf{0.322(1)}& \textbf{0.369(1)}& \textbf{0.903(1)}& 0.856(2)          & 0.314(2)          & 0.539(2)          & 0.649(2)          & \textbf{0.604(1)}\\ \hline
			\multirow{6}{*}{\textbf{APES\_local}\cite{APES2023cvpr}}  & \textbf{ST}        & 0.757(3)          & 0.708(4)          & 0.78(4)           & 0.847(4)          & 0.873(4)          & 0.87(4)           & 0.879(4)          & 0.831(4)          & 0.631(4)          & 0.42(2)           & 0.948(2)           & 0.921(6)          & 0.383(4)          & 0.7(4)            & 0.799(4)          & 0.756(4)          \\
			& \textbf{cutmix\_k\cite{Pointcutmix}}& 0.984(6)          & 1.011(5)          & 1.161(6)          & 1.082(6)          & 1.115(6)          & 1.217(6)          & 1.246(6)          & 1.128(6)          & 0.703(6)          & 1.066(6)          & 0.997(5)           & 0.869(4)          & 0.492(6)          & 0.807(6)          & 0.876(5)          & 0.984(6)          \\
			& \textbf{cutmix\_r\cite{Pointcutmix}} & 0.96(5)           & 1.048(6)          & 1.07(5)           & 1.005(5)          & 1.069(5)          & 1.044(5)          & 1.215(5)          & 0.974(5)          & 0.642(5)          & 0.982(5)          & 1.062(6)           & 0.88(5)           & 0.443(5)          & 0.732(5)          & 0.876(6)          & 0.933(5)          \\
			& \textbf{mixup\cite{Pointmixup}}     & \textbf{0.656(1)}& 0.645(2)          & 0.756(2)          & 0.736(2)          & 0.761(2)          & 0.757(2)          & 0.875(3)          & 0.668(2)          & 0.474(3)          & 0.734(4)          & 0.959(4)           & 0.854(3)          & 0.353(3)          & 0.617(3)          & 0.655(2)          & 0.7(3)            \\
			& \textbf{rsmix\cite{RSMix}}      & 0.848(4)          & 0.675(3)          & 0.758(3)          & 0.745(3)          & 0.774(3)          & 0.862(3)          & 0.843(2)          & 0.809(3)          & 0.43(2)           & 0.446(3)          & 0.957(3)           & \textbf{0.835(1)}& \textbf{0.303(1)}& \textbf{0.467(1)}& \textbf{0.536(1)}& 0.686(2)          \\
			& \textbf{DesenAT-sD(ours)}& 0.709(2)          & \textbf{0.52(1)}& \textbf{0.718(1)}& \textbf{0.696(1)}& \textbf{0.716(1)}& \textbf{0.65(1)}& \textbf{0.813(1)}& \textbf{0.576(1)}& \textbf{0.373(1)}& \textbf{0.391(1)}& \textbf{0.925(1)}& 0.853(2)          & 0.339(2)          & 0.573(2)          & 0.674(3)          & \textbf{0.635(1)}\\ \hline
		\end{tabular}
	}
\end{table}

% 7 CE
\begin{table}[!th]
	\centering
	\caption{CE of different model architectures on PointCloud-C with different training strategies.Bold: best in column. Underline: second best in column. \ddag: multi-scale inference from \cite{RepSurf}, ST: standard training.}\label{tab 4}
	\resizebox{\linewidth}{!}{%
		\begin{tabular}{llrrrrrrrr}\hline
			&                     & \textbf{add\_global}& \textbf{add\_local}& \textbf{dropout\_global}& \textbf{dropout\_local}& \textbf{jitter}& \textbf{rotate}& \textbf{scale}& \textbf{m\_CE}\\ \hline
			\multirow{6}{*}{\textbf{PointNet++\ddag}\cite{Pointnet++}}& \textbf{ST}       & 1.0(6)               & 1.0(6)              & 1.0(5)                   & 1.0(6)                  & 1.0(4)          & 1.0(2)          & 1.0(6)         & 1.0(5)         \\
			& \textbf{cutmix\_k\cite{Pointcutmix}}& 0.921(4)             & 0.896(4)            & 0.592(3)                 & 0.458(2)                & 1.019(5)        & 1.09(4)         & 0.953(4)       & 0.847(3)       \\
			& \textbf{cutmix\_r\cite{Pointcutmix}}& 0.961(5)             & 0.942(5)            & 0.373(2)                 & 0.629(3)                & 0.598(2)        & 1.129(5)        & 0.983(5)       & 0.802(2)       \\
			& \textbf{mixup\cite{Pointmixup}}    & 0.893(2)             & 0.875(3)            & 1.022(6)                 & 0.767(5)                & 0.945(3)        & 1.05(3)         & 0.832(3)       & 0.912(4)       \\
			& \textbf{rsmix\cite{RSMix}}     & \textbf{0.89(1)}            & \textbf{0.829(1)}          & 0.736(4)                 & \textbf{0.454(1)}              & 1.973(6)        & 1.552(6)        & 0.796(2)       & 1.033(6)       \\
			& \textbf{DesenAT-sD(ours)}& 0.919(3)             & 0.859(2)            &  \textbf{0.368(1)}               & 0.642(4)                & \textbf{0.385(1)}      & \textbf{0.862(1)}      & \textbf{0.616(1)}     & \textbf{0.664(1)}     \\ \hline
			\multirow{6}{*}{\textbf{PointNetMeta-S}\cite{pointNetMetaBase2023cvpr}}& \textbf{ST}       & 1.009(6)             & 0.941(6)            & 1.208(6)                 & 0.591(6)                & 1.081(4)        & 1.006(2)        & 1.022(6)       & 0.98(6)        \\
			& \textbf{cutmix\_k\cite{Pointcutmix}}&  \textbf{0.798(1)}           & \textbf{0.79(1)}           & 0.839(4)                 & 0.278(2)                & 1.783(5)        & 1.195(5)        & 0.864(4)       & 0.935(4)       \\
			& \textbf{cutmix\_r\cite{Pointcutmix}}& 0.882(5)             & 0.814(3)            & 0.516(2)                 & 0.383(3)                & 0.901(3)        & 1.12(4)         & 0.918(5)       & 0.791(2)       \\
			& \textbf{mixup\cite{Pointmixup}}    & 0.863(4)             & 0.847(5)            & 0.932(5)                 & 0.432(4)                & 0.819(2)        & 1.033(3)        & 0.794(2)       & 0.817(3)       \\
			& \textbf{rsmix\cite{RSMix}}     & 0.841(2)             & 0.809(2)            & 0.65(3)                  & \textbf{0.241(1)}              & 2.029(6)        & 1.4(6)          & 0.806(3)       & 0.968(5)       \\
			& \textbf{DesenAT-sD(ours)}& 0.86(3)              & 0.838(4)            & \textbf{0.477(1)}               & 0.478(5)                &  \textbf{0.359(1)}      & \textbf{0.953(1)}      & \textbf{0.525(1)}     & \textbf{0.641(1)}     \\ \hline
			\multirow{6}{*}{\textbf{APES\_global}\cite{APES2023cvpr}} & \textbf{ST}       & 0.784(2)             & 0.756(2)            & 0.706(4)                 & 0.457(4)                & 0.87(4)         & 0.802(3)        & 0.811(2)       & 0.741(4)       \\ 
			& \textbf{cutmix\_k\cite{Pointcutmix}}& 0.956(5)             & 0.923(5)            & 0.929(6)                 & 0.4(2)                  & 0.902(5)        & 0.912(6)        & 0.974(6)       & 0.857(6)       \\
			& \textbf{cutmix\_r\cite{Pointcutmix}}& 0.86(4)              & 0.861(4)            & 0.388(2)                 & 0.499(5)                & 0.583(2)        & 0.871(4)        & 0.916(4)       & 0.711(3)       \\
			& \textbf{mixup\cite{Pointmixup}}    & 1.0(6)               & 0.971(6)            & 0.872(5)                 & 0.533(6)                & 0.708(3)        & 0.762(2)        & 0.945(5)       & 0.827(5)       \\
			& \textbf{rsmix\cite{RSMix}}     & 0.835(3)             & 0.799(3)            & \textbf{0.359(1)}               & \textbf{0.256(1)}              & 0.911(6)        & 0.875(5)        & 0.834(3)       & 0.696(2)       \\
			& \textbf{DesenAT-sD(ours)}& \textbf{0.78(1)}            & \textbf{0.736(1)}          & 0.427(3)                 & 0.421(3)                & \textbf{0.347(1)}      & \textbf{0.69(1)}       & \textbf{0.519(1)}     & \textbf{0.56(1)}      \\ \hline
			\multirow{6}{*}{\textbf{APES\_local}\cite{APES2023cvpr}}  & \textbf{ST}       & 0.855(3)             & 0.817(3)            & 0.692(5)                 & 0.455(4)                & 0.796(4)        & 0.762(3)        & 0.779(2)       & 0.737(4)       \\ 
			& \textbf{cutmix\_k\cite{Pointcutmix}}& 1.225(6)             & 1.206(5)            & 0.948(6)                 & 0.485(5)                & 1.037(6)        & 0.988(6)        & 1.152(6)       & 1.006(6)       \\
			& \textbf{cutmix\_r\cite{Pointcutmix}}& 1.191(5)             & 1.212(6)            & 0.638(3)                 & 0.572(6)                & 0.776(3)        & 0.945(5)        & 1.123(5)       & 0.922(5)       \\
			& \textbf{mixup\cite{Pointmixup}}    & 0.862(4)             & 0.835(4)            & 0.648(4)                 & 0.413(2)                & 0.512(2)        & \textbf{0.675(1)}      & 0.824(3)       & 0.681(2)       \\
			& \textbf{rsmix\cite{RSMix}}     & 0.836(2)             & 0.789(2)            & \textbf{0.382(1)}               & \textbf{0.266(1)}              & 0.808(5)        & 0.894(4)        & 0.832(4)       & 0.687(3)       \\
			& \textbf{DesenAT-sD(ours)}& \textbf{0.795(1)}           & \textbf{0.754(1)}          & 0.503(2)                 & 0.422(3)                & \textbf{0.404(1)}      & 0.749(2)        & \textbf{0.593(1)}     & \textbf{0.603(1)}     \\ \hline
		\end{tabular}
	}
\end{table}

%\begin{figure}[t]
%	\centering
%	\includegraphics[width=0.8\linewidth]{Visualization.eps}
%	\caption{(a) Visualization of the PointNet++ model's sensitivity to point cloud samples trained by the ST method. (b) Histogram of the PointNet++ model's sensitivity distribution to point cloud samples trained by the ST method. (c) Visualization of the PointNet++ model's sensitivity to point cloud samples trained by the DesenAT method. (d) Histogram of the PointNet++ model's sensitivity distribution to point cloud samples trained by the DesenAT method.}
%	\label{Visualization}
%\end{figure}

In Tabs. \ref{tab 3} and \ref{tab 4}, the values in parentheses represent the rankings of each method among the six approaches. The tables show that the proposed method (DesenAT-sD) generally performed, with m\_CE consistently ranking first. This confirms the competitiveness and superiority of the proposed approach. Furthermore, the rankings of DesenAT-sD across various metrics demonstrated their versatility. Across all indicators, DesenAT-sD consistently maintained a high ranking with minimal fluctuations. For example, considering the PointNetMeta-S model on the ModelNet-C dataset, Cutmix-r attains the first rank in five benchmarks (Ups., Impulse, LiDAR, Inc., Dec.); however, it drops to sixth place in the Occlusion benchmark. In contrast, DesenAT-sD consistently maintained the top four rankings across all metrics. Similarly, on the PointCloud-C dataset employing PointNet++(MSG) as an example, Rsmix achieved the first rank in the add\_global, add\_local, and dropout\_local benchmarks; however, it ranked sixth in the jitter and rotation benchmarks. In contrast, DesenAT-sD secured the first rank in add\_global, jitter, rotation, and scale benchmarks, with a fourth place ranking in the dropout\_local benchmark. This indicates that methods such as Cutmix-k, Cutout, Mixup, and Rsmix are effective only in specific corruption scenarios and lack general applicability. In contrast, DesenAT-sD consistently demonstrated an impressive performance across all indicators.

% Clean
\begin{table}[!th]
	\centering
	\caption{OA\% with different training strategies on the clean ModelNet40 dataset. Bold: best in column. \ddag: multi-scale inference from \cite{RepSurf}, ST: standard training. }\label{tab 5}
	\resizebox{\linewidth}{!}{%
		\begin{tabular}{lrrrr}\hline
			& \multicolumn{1}{c}{\textbf{PointNet++\ddag}\cite{Pointnet++}}& \textbf{PointNetMeta-S}\cite{pointNetMetaBase2023cvpr}& \textbf{APES\_global}\cite{APES2023cvpr}& \textbf{APES\_local}\cite{APES2023cvpr}\\ \hline
			\textbf{ST}                               & 91.25                          & 91.73          & 92.67          & 93.23          \\
			\textbf{cutmix\_k\cite{Pointcutmix}}      & 92.34                          & \textbf{93.15} & 89.51          & 91.69          \\
			\textbf{cutmix\_r\cite{Pointcutmix}}      & 91.86                          & 92.54          & 89.59          & 92.54          \\
			\textbf{mixup\cite{Pointmixup}}           & 92.34                          & 92.54          & 92.54          & 91.57          \\
			\textbf{rsmix\cite{RSMix}}                & \textbf{92.38}                 & 92.75          & 92.91           & 92.87          \\
			\textbf{DesenAT-sD(ours)}                 & 92.14                          & 92.46          & \textbf{93.11} & \textbf{93.35}\\ \hline
		\end{tabular}
	}
\end{table}

Notably, as shown in Tab. \ref{tab 5}, the proposed approach enhances robustness without compromising the performance of the model on clean datasets.

%Additionally, we provide a visualization of the sensitivity of models trained with the ST method and the DesenAT-sD method to the original and corrupted samples. In Fig. \ref{Visualization}, by comparing (a) and (c), it is visually evident that the model trained with the ST method is more sensitive to certain feature points in the point cloud sample (the redder points in the figure), whereas the model trained with the DesenAT method exhibits a more uniform sensitivity to these points. This can also be intuitively seen from (b) and (d). In the histogram of (b), there are a few points distributed near the right side (high sensitivity value region), while in (d), there are no points in the high sensitivity region. This indicates that the method proposed in this paper (DesenAT) can indeed smooth the model's sensitivity to the samples.

\subsection{Ablation Studies}\label{s4.3}
The ablation experiments consisted of two steps. In Sec. \ref{s4.3.1}, we designed comparative experiments for the DesenAT framework (Sec. \ref{s4.3.2}) and the DesenAT-sD framework (Sec. \ref{s4.3.1}), aiming to verify the effectiveness of DesenAT and the distillation algorithm. Subsequently, we broke down the DesenAT framework into two key steps: `Shapley Filtration' and `spatial transformation' and validated the effectiveness of each step individually.
\subsubsection{Experiments on Adversarial Samples}\label{s4.3.1}
% As1 OA
\begin{table}[h]
	\centering
	\caption{The mean OA(\%) under 5 corruption levels of different model architectures on ModelNet40-C with DesenAT and DesenAT-SD}\label{tab 6}
	\resizebox{\linewidth}{!}{%
		\begin{tabular}{rrllllllllllllllll} \hline
			\multicolumn{1}{l}{}                                                   & \multicolumn{1}{l}{}      & \multicolumn{1}{c}{\textbf{rotation}} & \multicolumn{1}{c}{\textbf{shear}} & \multicolumn{1}{c}{\textbf{distortion}} & \multicolumn{1}{c}{\textbf{distortion\_rbf}} & \multicolumn{1}{c}{\textbf{distortion\_rbf\_inv}} & \multicolumn{1}{c}{\textbf{uniform}} & \multicolumn{1}{c}{\textbf{upsampling}} & \multicolumn{1}{c}{\textbf{gaussian}} & \multicolumn{1}{c}{\textbf{impulse}} & \multicolumn{1}{c}{\textbf{background}} & \multicolumn{1}{c}{\textbf{occlusion}} & \multicolumn{1}{c}{\textbf{lidar}} & \multicolumn{1}{c}{\textbf{density\_inc}} & \multicolumn{1}{c}{\textbf{density}} & \multicolumn{1}{c}{\textbf{cutout}} & \multicolumn{1}{c}{\textbf{avg}} \\ \hline
			\multirow{5}{*}{\textbf{PointNet++\ddag\cite{Pointnet++}}} & \textbf{ST}               & 80.09                                 & 84.90                              & 86.54                                   & 86.43                                        & 87.51                                             & 87.67                                & 90.67                                   & 85.61                                 & 72.60                                & 77.95                                   & 41.01                                  & 31.78                              & 71.59                                     & 80.58                                & 83.74                               & 76.58                            \\
			& \textbf{DesenAT}          & \textbf{88.07}                        & 89.96                              & 89.13                                   & 89.98                                        & 90.17                                             & 91.14                                & 91.18                                   & 91.06                                 & 89.40                                & 77.52                                   & 45.75                                  & 46.39                              & 87.63                                     & 89.26                                & 89.84                               & 83.10                            \\
			& \textbf{DesenAT+DLB\cite{shen2022self}}      & 83.12                                 & \textbf{91.76}                     & \textbf{89.46}                          & 89.55                                        & 90.16                                             & 91.54                                & \textbf{91.99}                          & 91.56                                 & \textbf{91.21}                       & \textbf{91.96}                          & 52.76                                  & 38.97                              & \textbf{91.10}                            & \textbf{91.13}                       & \textbf{90.95}                      & 84.48                            \\
			& \textbf{DesenAT+PSKD\cite{kim2021self}}     & 83.52                                 & 90.97                              & 88.43                                   & 88.67                                        & 89.38                                             & 90.68                                & 90.84                                   & 90.56                                 & 90.59                                & 91.86                                   & \textbf{53.59}                         & \textbf{57.02}                     & 90.24                                     & 90.01                                & 89.94                               & 85.09                            \\
			& \textbf{DesenAT-sD(ours)} & 84.98                                 & 91.62                              & 89.13                                   & \textbf{90.05}                               & \textbf{90.45}                                    & \textbf{91.64}                       & 91.92                                   & \textbf{91.57}                        & 90.92                                & 88.53                                   & 52.36                                  & 54.12                              & 89.20                                     & 90.40                                & 90.21                               & \textbf{85.14} \\ \hline                  
		\end{tabular}
	}
\end{table}

% As1 CE
\begin{table}[h]
	\centering
	\caption{CE(\%) under 5 corruption levels of different model architectures on ModelNet40-C with DesenAT and DesenAT-SD}\label{tab 7}
	\resizebox{\linewidth}{!}{%
		\begin{tabular}{llrrrrrrrrrrrrrrrr} \hline  
			&                           & \textbf{Rotation} & \textbf{Shear} & \textbf{FFD} & \textbf{RBF} & \textbf{Lnv.RBF} & \textbf{Uniform} & \textbf{Ups.} & \textbf{Gaussian} & \textbf{Impulse} & \textbf{Bg.} & \textbf{Occlusion} & \textbf{LiDAR} & \textbf{Inc.} & \textbf{Dec.} & \textbf{Cutout} & \textbf{m\_CE} \\ \hline  
			\multirow{5}{*}{\textbf{PointNet++\ddag\cite{Pointnet++}}} & \textbf{ST}               & 1.0               & 1.0            & 1.0          & 1.0          & 1.0              & 1.0              & 1.0           & 1.0               & 1.0              & 1.0          & 1.0                & 1.0            & 1.0           & 1.0           & 1.0             & 1.0            \\
			& \textbf{DesenAT}          & 0.599             & 0.665          & 0.808        & 0.738        & 0.787            & 0.718            & 0.945         & 0.621             & 0.387            & 1.02         & 0.92               & 0.786          & 0.436         & 0.553         & 0.625           & 0.707          \\
			& \textbf{DesenAT+DLB}      & 0.848             & 0.546          & 0.783        & 0.77         & 0.788            & 0.686            & 0.859         & 0.586             & 0.321            & 0.365        & 0.801              & 0.895          & 0.313         & 0.457         & 0.557           & 0.638          \\
			& \textbf{DesenAT+PSKD}     & 0.828             & 0.598          & 0.86         & 0.835        & 0.85             & 0.756            & 0.982         & 0.656             & 0.344            & 0.369        & 0.787              & 0.63           & 0.343         & 0.514         & 0.619           & 0.665          \\
			& \textbf{DesenAT-sD(ours)} & 0.755             & 0.555          & 0.808        & 0.733        & 0.765            & 0.677            & 0.866         & 0.586             & 0.331            & 0.52         & 0.808              & 0.673          & 0.38          & 0.494         & 0.602           & 0.637     \\ \hline      
		\end{tabular}
	}
\end{table}

We compared our method with current mainstream self-distillation approaches and incorporated the corresponding code into our work. Notably, both DLB and PSKD distill knowledge from the previous batch of data to enhance the learning of the current batch. In contrast, our method learns knowledge within the same batch, which is fundamentally different.

To verify the compatibility of the DesenAT method, we followed the practices in \cite{shen2022self} and \cite{kim2021self}, enabling the model to learn from the previous batch of data to achieve consistency regularization. In the DesenAT+DLB and DesenAT-PSKD experiments, only adversarial samples, without original samples, were used as input during the comparison.

The results in Tabs. \ref{tab 6} and \ref{tab 7} demonstrate that DesenAT is not only compatible with self-distillation methods that learn within the same batch but also with methods that learn across different batches, such as DLB and PSKD.

\subsubsection{Blation Experiments on Adversarial Samples}\label{s4.3.2}

% As2 OA
\begin{table}[h]
	\centering
	\caption{The mean OA(\%) under 5 corruption levels of PointNet++ model on ModelNet40-C with different training strategies. \ddag: multi-scale inference from \cite{RepSurf}, ST: standard training, RF: Random Filtration, SF: Shapley Filtration, spatialT: spatial transformation.}\label{tab 8}
	\resizebox{\linewidth}{!}{%
		\begin{tabular}{llrrrrrrrrrrrrrrrr}\hline
			\multicolumn{1}{l}{}               & \multicolumn{1}{l}{}& \multicolumn{1}{c}{\textbf{Rotation}}& \multicolumn{1}{c}{\textbf{Shear}}& \multicolumn{1}{c}{\textbf{FFD}}& \multicolumn{1}{c}{\textbf{RBF}}& \multicolumn{1}{c}{\textbf{Lnv.RBF}}& \multicolumn{1}{c}{\textbf{Uniform}}& \multicolumn{1}{c}{\textbf{Ups.}}& \multicolumn{1}{c}{\textbf{Gaussian}}& \multicolumn{1}{c}{\textbf{Impulse}}& \multicolumn{1}{c}{\textbf{Bg.}}& \multicolumn{1}{c}{\textbf{Occlusion}}& \multicolumn{1}{c}{\textbf{LiDAR}}& \multicolumn{1}{c}{\textbf{Inc.}}& \multicolumn{1}{c}{\textbf{Dec.}}& \multicolumn{1}{c}{\textbf{Cutout}}& \multicolumn{1}{c}{\textbf{mean\_OA}}\\ \hline
			\multirow{5}{*}{\textbf{PointNet++\ddag}\cite{Pointnet++}}& \textbf{ST}        & 80.09                                 & 84.9                               & 86.54                            & 86.43                            & 87.51                                & 87.67                                & 90.67                             & 85.61                                 & 72.6                                 & 77.95                            & 41.01                                  & 31.78                              & 71.59                             & 80.58                             & 83.74                               & 76.58                                 \\
			& \textbf{RF}        & 79.92                        & 84.35                              & 86.32                            & 86.64                            & 86.82                                & 87.57                                & 91                                & 84.52                                 & 71.46                                & 77.42                            & 42.19                                  & 33.42                              & 70.65                             & 84.51                             & 85.87                               & 76.84                                 \\
			& \textbf{SF}        & 80.17                                 & 85.81                     & 87.01                   & 87.67                   & 88.28                       & 88.65                       & \textbf{91.47}                  & 86.37                        & 80                          & \textbf{79.97}                 & 44.58                         & 42.05                     & 85.43                    & \textbf{89.56}                  & \textbf{90.23}                    & 80.48                        \\
			& \textbf{spatialT}  & \textbf{88.18}                      & \textbf{90.09}                   & 89.11                            & 89.92                            & 89.97                                & \textbf{91.28}                     & 90.84                             & \textbf{91.15}                      & \textbf{89.6}                      & 75.18                            & 39.85                                  & 31.38                              & 78.43                             & 83.7                              & 86.21                               & 80.33                                 \\
			& \textbf{DesenAT}   & 88.07                                 & 89.96                     & \textbf{89.13}                 & \textbf{89.98}                 & \textbf{90.17}                     & 91.14                                & 91.18                    & 91.06                                 & 89.4                                 & 77.52                   & \textbf{45.75}                       & \textbf{46.39}                   & \textbf{87.63}                  & 89.26                    & 89.84                      & \textbf{83.1}                       \\ \hline
		\end{tabular}
	}
\end{table}

% As2 CE
\begin{table}[h]
	\centering
	\caption{CE(\%) under 5 corruption levels of PointNet++ model on ModelNet40-C with different training strategies. \ddag: multi-scale inference from \cite{RepSurf}, ST: standard training, RF: Random Filtration, SF: Shapley Filtration, spatialT: spatial transformation. }\label{tab 9}
	\resizebox{\linewidth}{!}{%
		\begin{tabular}{llrrrrrrrrrrrrrrrr}\hline
			&                   & \multicolumn{1}{c}{\textbf{Rotation}}& \multicolumn{1}{c}{\textbf{Shear}}& \multicolumn{1}{c}{\textbf{FFD}}& \multicolumn{1}{c}{\textbf{RBF}}& \multicolumn{1}{c}{\textbf{Lnv.RBF}}& \multicolumn{1}{c}{\textbf{Uniform}}& \multicolumn{1}{c}{\textbf{Ups.}}& \multicolumn{1}{c}{\textbf{Gaussian}}& \multicolumn{1}{c}{\textbf{Impulse}}& \multicolumn{1}{c}{\textbf{Bg.}}& \multicolumn{1}{c}{\textbf{Occlusion}}& \multicolumn{1}{c}{\textbf{LiDAR}}& \multicolumn{1}{c}{\textbf{Inc.}}& \multicolumn{1}{c}{\textbf{Dec.}}& \multicolumn{1}{c}{\textbf{Cutout}}& \multicolumn{1}{c}{\textbf{m\_CE}}\\ \hline
			\multirow{5}{*}{\textbf{PointNet++\ddag}\cite{Pointnet++}}& \textbf{ST}     & 1                                     & 1                                  & 1                                & 1                                & 1                                    & 1                                    & 1                                 & 1                                     & 1                                    & 1                                & 1                                      & 1                                  & 1                                 & 1                                 & 1                                   & 1                                  \\
			& \textbf{RF}     & 1.009                                 & 1.036                              & 1.016                            & 0.985                            & 1.056                                & 1.008                                & 0.965                             & 1.075                                 & 1.042                                & 1.024                            & 0.98                                   & 0.976                              & 1.033                             & 0.797                             & 0.869                               & 0.991                              \\
			& \textbf{SF}     & 0.996                                 & 0.94                               & 0.965                            & 0.908                            & 0.938                                & 0.92                                 & \textbf{0.915}                  & 0.947                                 & 0.73                                 & \textbf{0.908}                 & 0.939                                  & 0.85                               & 0.513                             & \textbf{0.537}                  & \textbf{0.601}                    & 0.84                               \\
			& \textbf{spatialT}& \textbf{0.594}                      & 0.656                              & 0.809                            & 0.743                            & 0.803                                & \textbf{0.707}                     & 0.982                             & \textbf{0.615}                      & \textbf{0.38}                      & 1.126                            & 1.02                                   & 1.006                              & 0.759                             & 0.839                             & 0.848                               & 0.792                              \\
			& \textbf{DesenAT}& 0.599                                 & \textbf{0.665}                   & \textbf{0.808}                 & \textbf{0.738}                 & \textbf{0.787}                     & 0.718                                & 0.945                             & 0.621                                 & 0.387                                & \textbf{1.02}                  & \textbf{0.92}                        & \textbf{0.786}                   & \textbf{0.436}                  & 0.553                             & 0.625                               & \textbf{0.707}                   \\ \hline
		\end{tabular}
	}
\end{table}

As shown in Tabs. \ref{tab 8} and \ref{tab 9}, we first compared RF and SF, using the ST method as a baseline. The random filtering (RF) method did not show a substantial improvement in overall accuracy, indicating that randomly removing points from the point cloud does not enhance robustness. In contrast, the Shapley filtering (SF) method significantly improved the model robustness after removing the sensitive points (compared to ST, m\_OA increased by 3.9 percentage points). This suggests that the proposed SF method can improve model robustness to a certain extent.

When only using the spatialT method for training, the model robustness showed some improvement over ST. We observed that the spatialT method performed better than SF on morphological corruption benchmarks (Rotation, Shear, FFD, RBF, Lnv.RBF), whereas SF outperformed spatialT on density corruption benchmarks (Occlusion, LiDAR, Inc., Dec, and Cutout). This is easy to understand: spatialT involves geometric transformations, which makes the model more robust to morphological corruption, whereas SF is a point removal operation, which is robust to density changes.

By combining the SF and spatialT methods, the m\_OA index improved by approximately 2.5 percentage points compared with using either method alone, demonstrating that the two methods have strong fault tolerance capabilities.

\section{Conclusion}\label{s5}
\textbf{Discussion:} We have observed that neural networks rely too heavily on certain features for classification, leading to model fragility. To address this, we introduce Shapley values to enhance point-cloud corruption robustness. Shapley values measure each feature's contribution to the model’s prediction, quantifying its sensitivity. We propose the DesenAT framework to eliminate highly sensitive features, forcing the model to focus on overlooked ones. This enables the model to recognize samples using alternative features even when key ones are corrupted.

\textbf{Limitation:} While DesenAT effectively improves robustness, calculating feature contributions using Shapley values is computationally feasible mainly for classification tasks with low complexity and limited output space. Tasks like segmentation and object detection, with larger output spaces and higher complexity, present challenges in assessing feature sensitivity.

\textbf{Prospects:} Attribution operations were conducted on point clouds in the spatial domain, with future research potentially exploring frequency or graph domains. By analyzing model sensitivity to frequency components, we can identify model vulnerabilities and propose training methods (e.g., adding noise to the frequency domain) to strengthen robustness against corruption.

We hope this study encourages researchers to rethink robustness tasks. Given the inevitable corruption in real-world point clouds, we believe feature desensitization is worth further exploration in downstream tasks like segmentation and object detection. Our approach may also extend to other machine learning areas, such as 2D image processing, to assess whether attribution methods can enhance 2D neural network robustness, considering the complexity of 2D feature spaces.

\section{Acknowledgment}\label{s6}
This work was partially supported by the National Natural Science Foundation of China under Grants (No.51774219), and the key R\&D Program of Hubei Province (No.2020BAB098).

Numerical calculation is supported by High-Performance Computing Center of Wuhan University of Science and Technology.

\bibliographystyle{plain}
\bibliography{ours.bib}

% \newpage
\appendix
\section{Experimental Results}
The appendix presents the performance of different models under various training methods at different corruption levels.

\begin{figure*}[h] 
	\centering
	\begin{minipage}{0.45\textwidth}
		\centering
		\includegraphics[width=\linewidth]{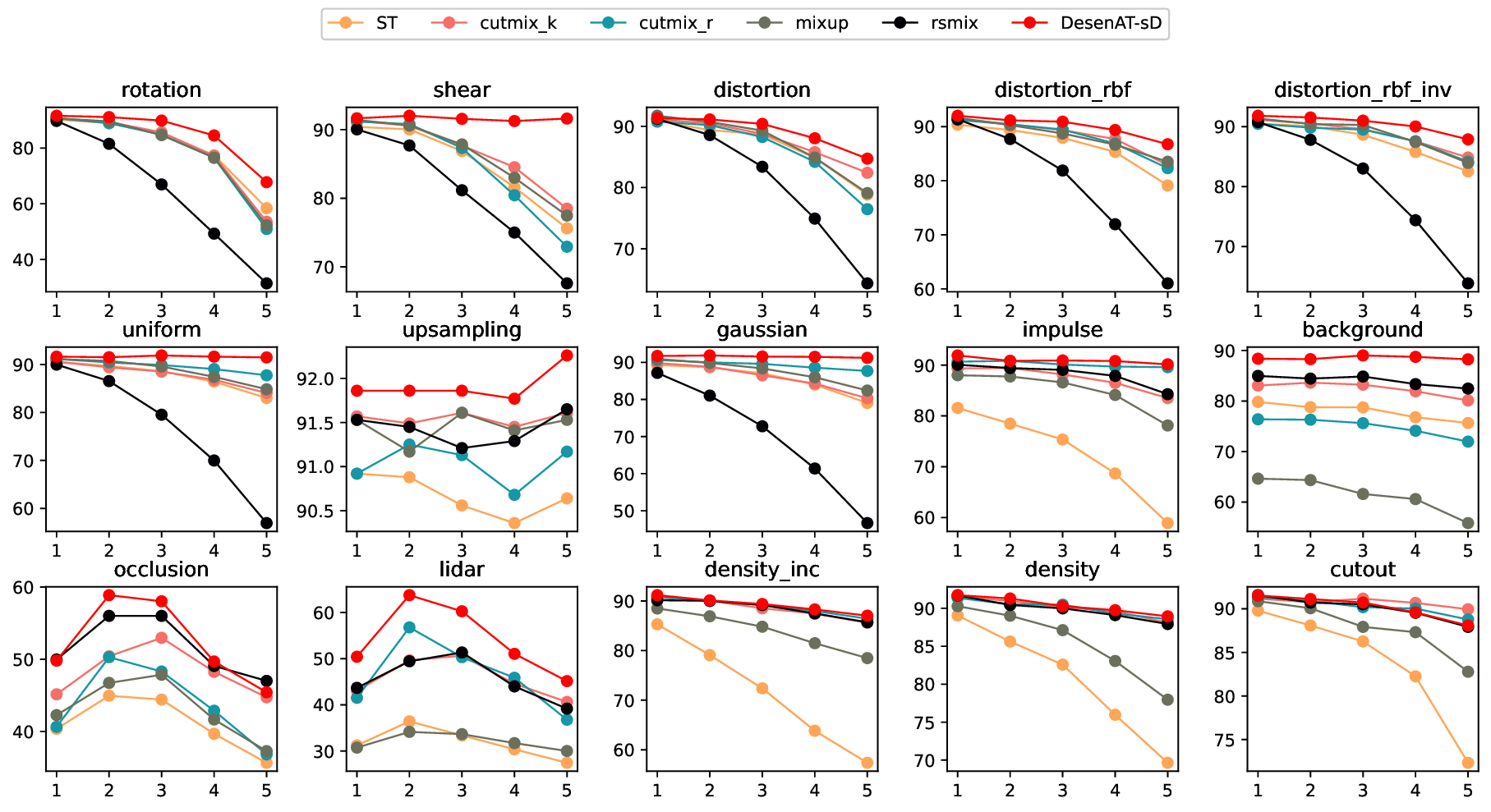}
		\caption{OA(\%) of PointNet++(msg) with different data augmentation strategies on modelNet40-C under different robustness levels.}
		\label{Fig7}
	\end{minipage}
	\hfill
	\begin{minipage}{0.45\textwidth}
		\centering
		\includegraphics[width=\linewidth]{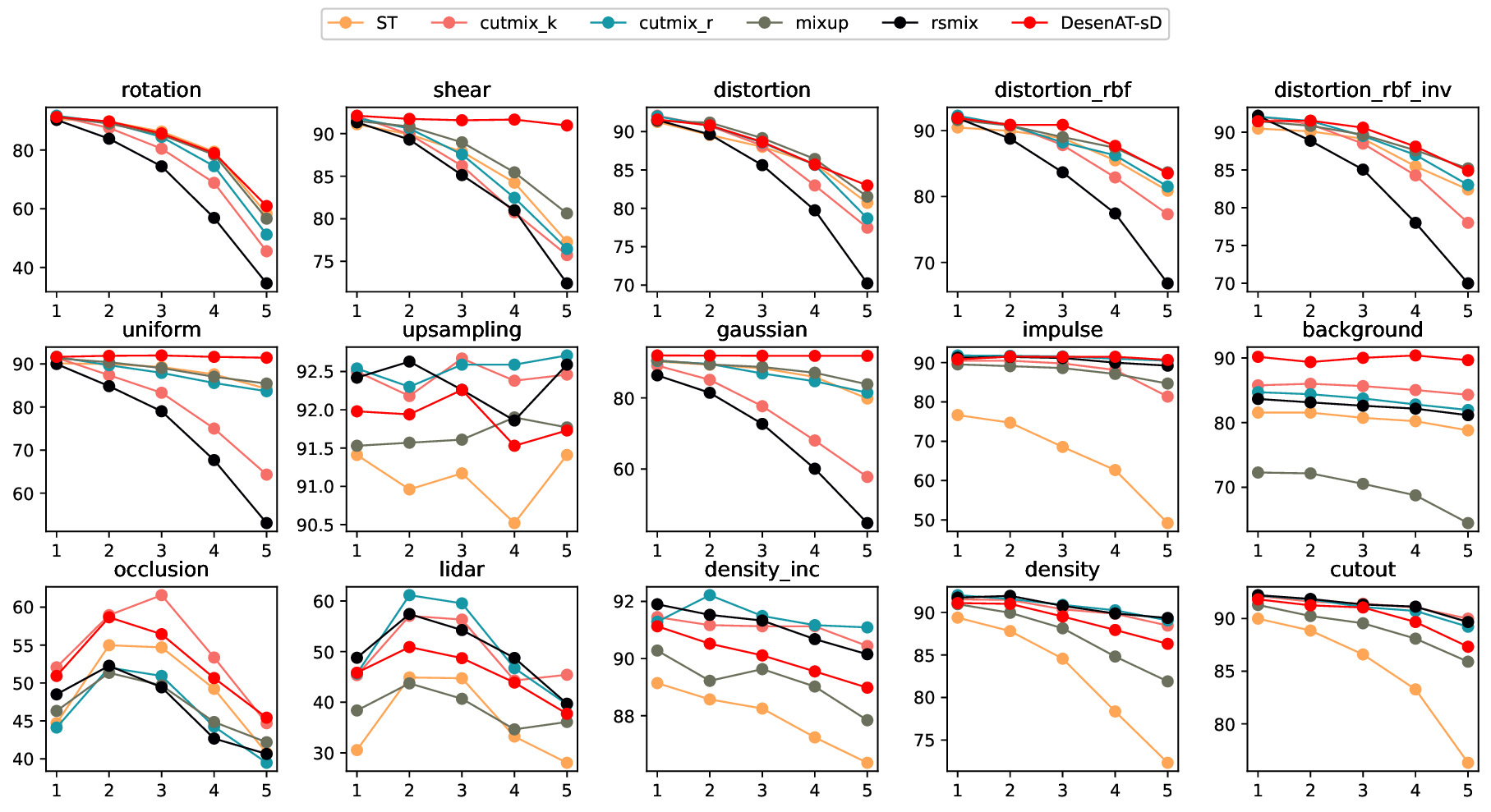}
		\caption{OA(\%) of PointNetMeta-S with different data augmentation strategies on modelNet40-C under different robustness levels.}
		\label{Fig8}
	\end{minipage}
	
	\vspace{0.5cm}
	
	\begin{minipage}{0.45\textwidth}
		\centering
		\includegraphics[width=\linewidth]{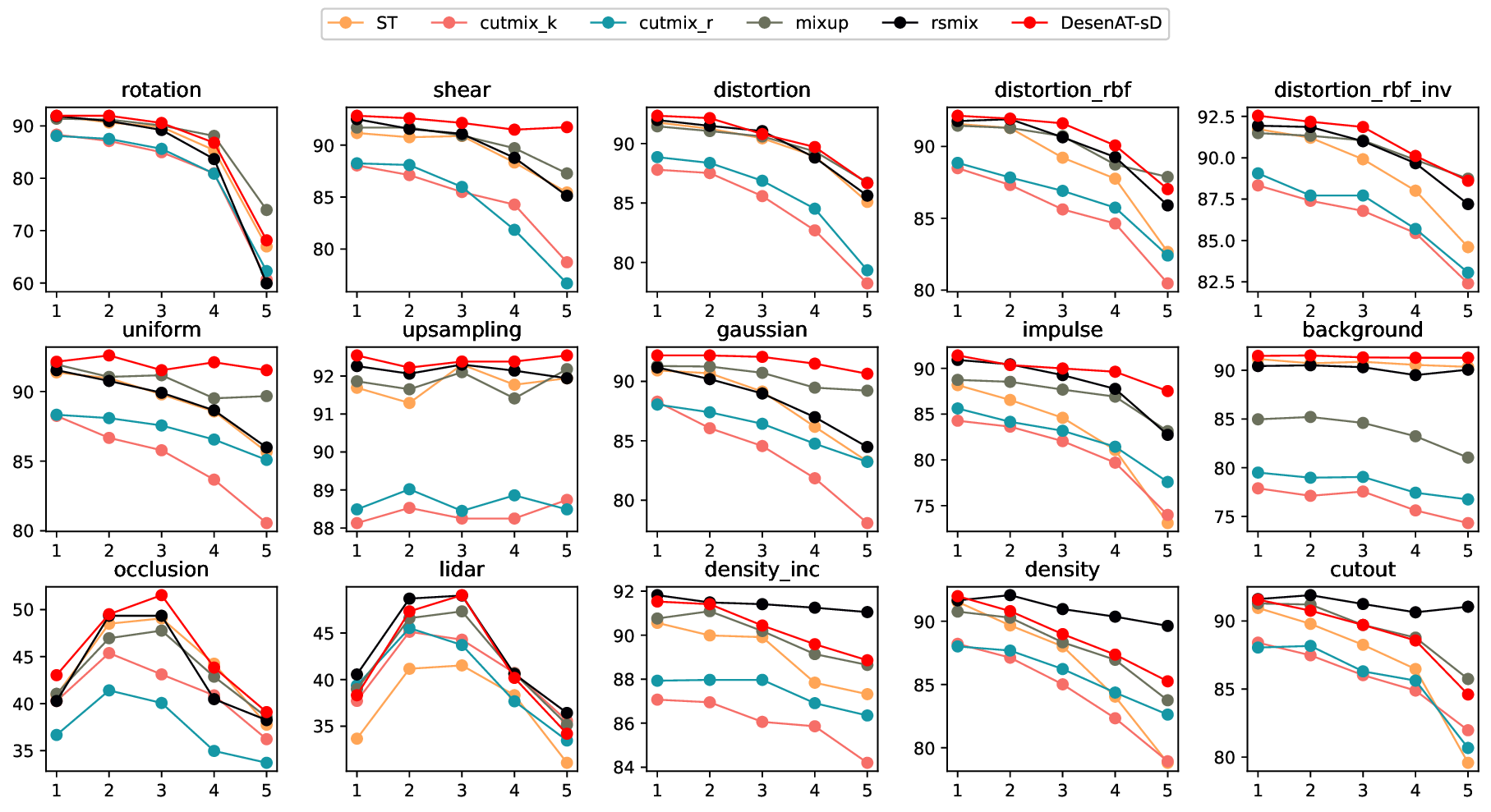}
		\caption{OA(\%) of APES\_local with different data augmentation strategies on modelNet40-C under different robustness levels.}
		\label{Fig9}
	\end{minipage}
	\hfill
	\begin{minipage}{0.45\textwidth}
		\centering
		\includegraphics[width=\linewidth]{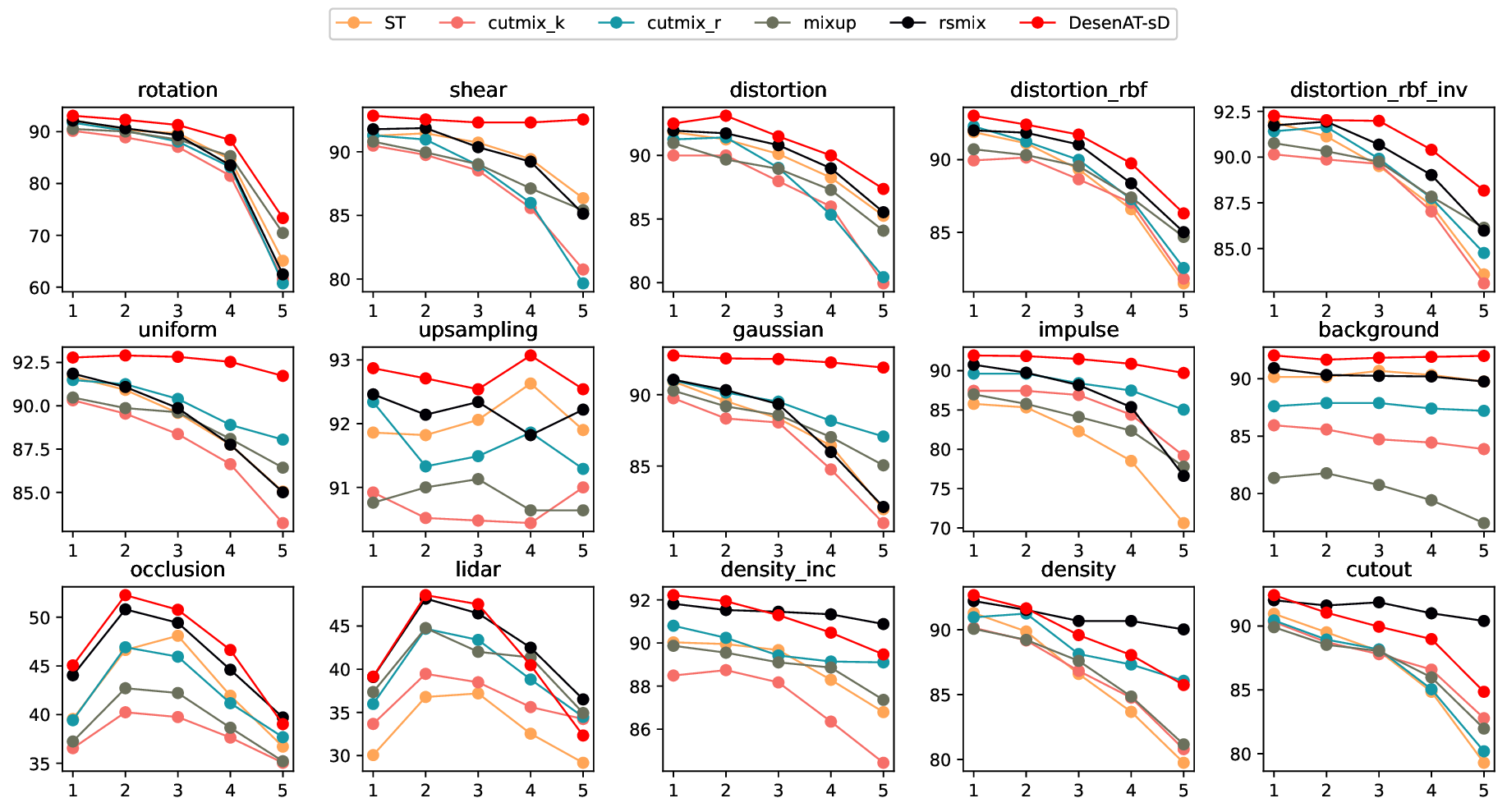}
		\caption{OA(\%) of APES\_global with different data augmentation strategies on modelNet40-C under different robustness levels.}
		\label{Fig10}
	\end{minipage}
	
	\vspace{0.5cm}
	
	\begin{minipage}{0.45\textwidth}
		\centering
		\includegraphics[width=\linewidth]{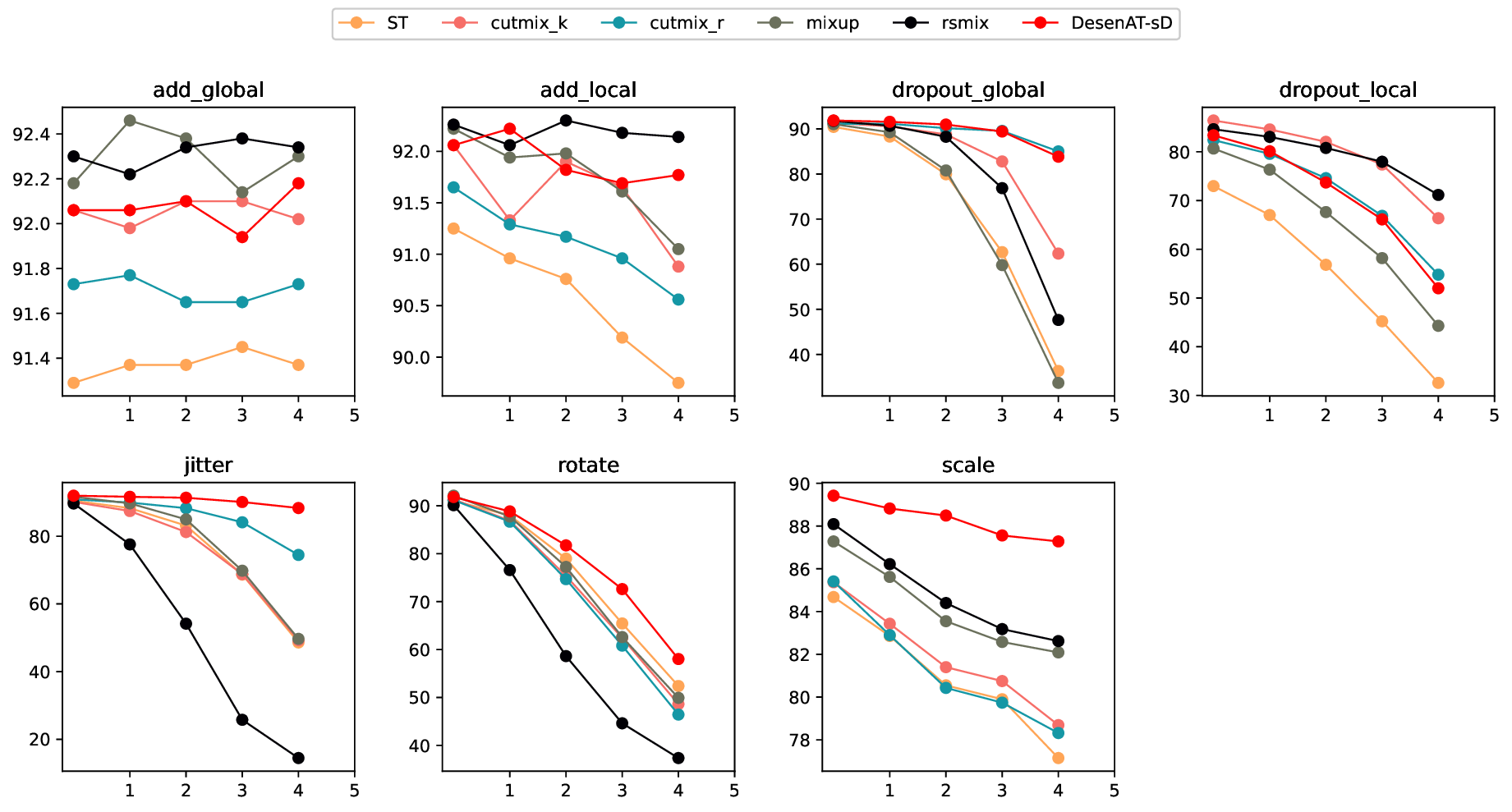}
		\caption{OA(\%) of PointNet++(msg) with different data augmentation strategies on Pointcloud-C under different robustness levels.}
		\label{Fig11}
	\end{minipage}
	\hfill
	\begin{minipage}{0.45\textwidth}
		\centering
		\includegraphics[width=\linewidth]{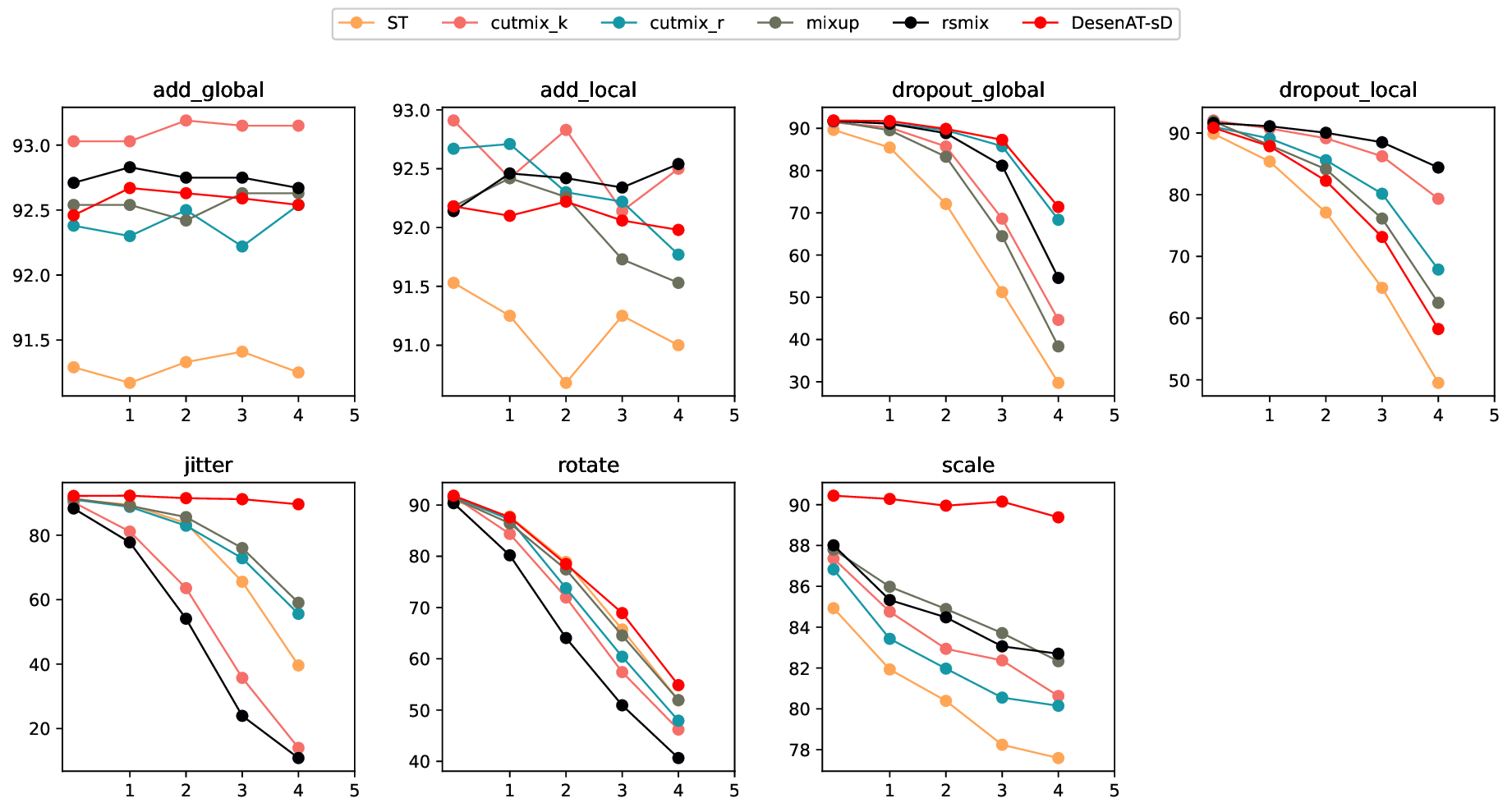}
		\caption{OA(\%) of PointNetMeta-S with different data augmentation strategies on Pointcloud-C under different robustness levels.}
		\label{Fig12}
	\end{minipage}
	
	\vspace{0.5cm}
	
	\begin{minipage}{0.45\textwidth}
		\centering
		\includegraphics[width=\linewidth]{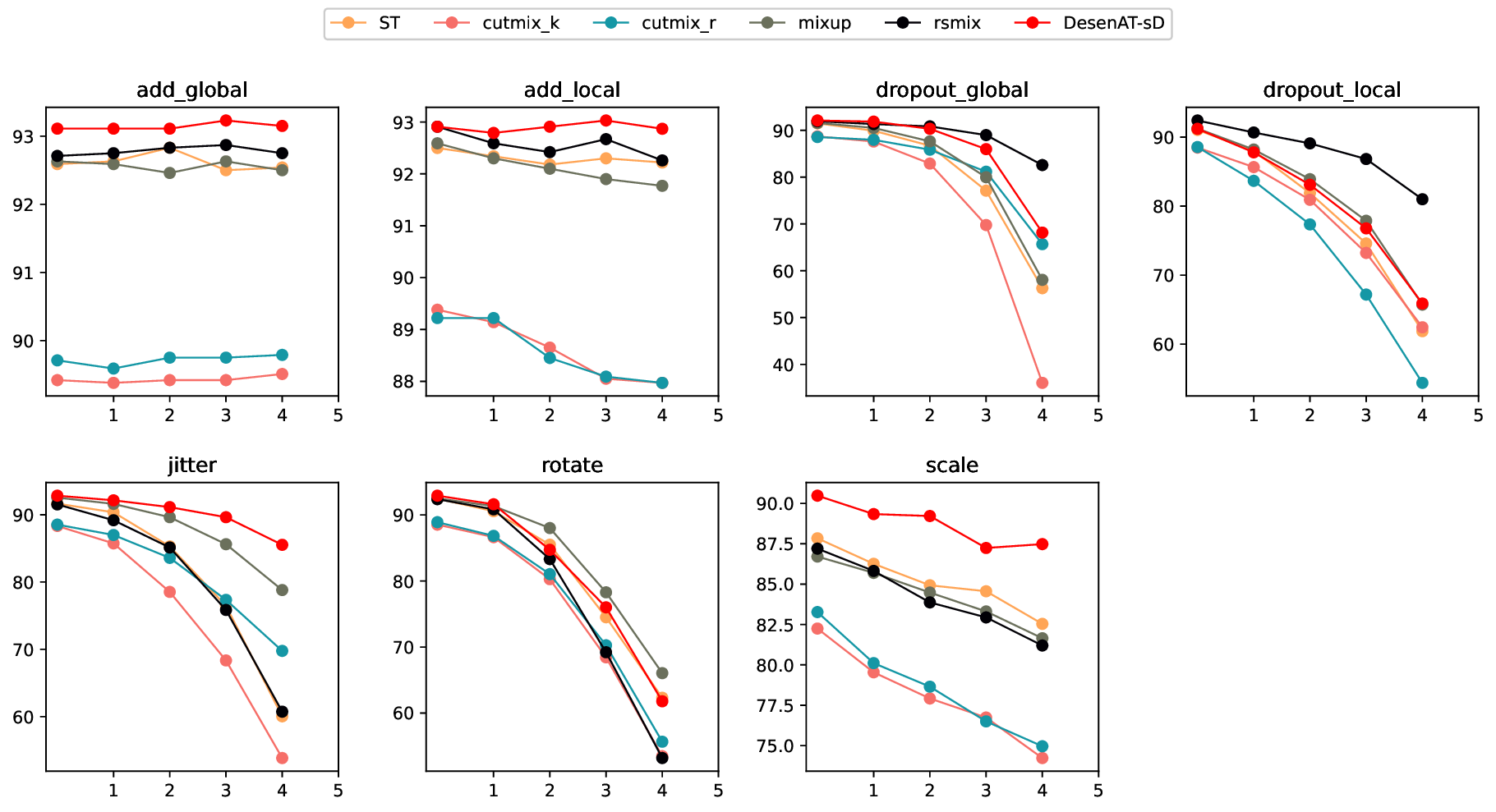}
		\caption{OA(\%) of APES\_local with different data augmentation strategies on Pointcloud-C under different robustness levels.}
		\label{Fig13}
	\end{minipage}
	\hfill
	\begin{minipage}{0.45\textwidth}
		\centering
		\includegraphics[width=\linewidth]{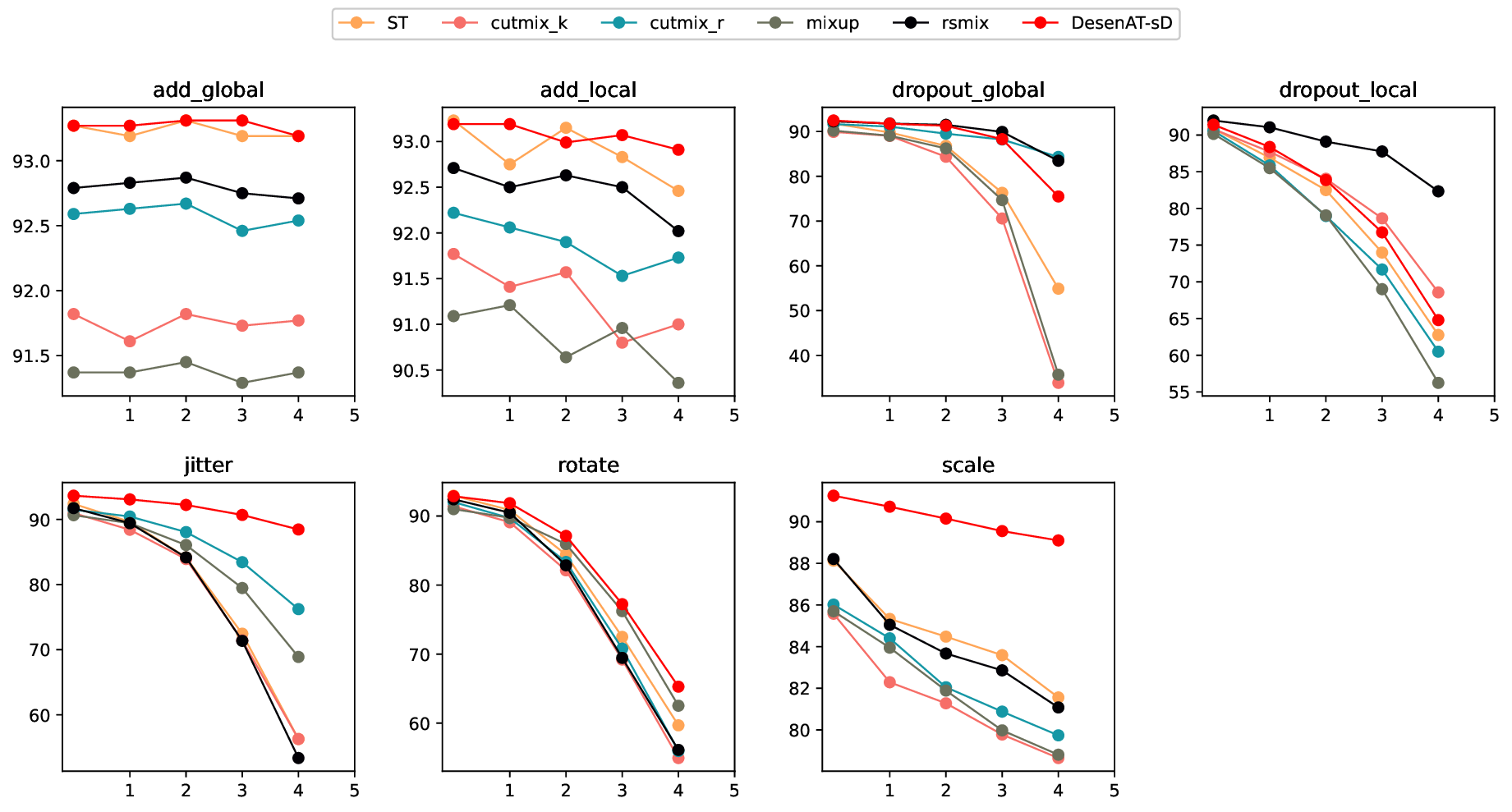}
		\caption{OA(\%) of APES\_global with different data augmentation strategies on Pointcloud-C under different robustness levels.}
		\label{Fig14}
	\end{minipage}
\end{figure*}

\clearpage
\section{Algorithm}
We wrote pseudo-code \ref{alg1} to describe the entire process of DesenAT: where steps 1-9 compute the Shapley values of the samples, corresponding to Eq. (3). Steps 10-12 remove points with high sensitivity based on the Shapley values, corresponding to the function $F(\cdot)$ in Eq. (8). The purpose of this is to force the model to learn from points with lower sensitivity during the subsequent training process. Steps 13-22 involve spatial transformations, corresponding to Eq. (7).

\begin{algorithm}
	\caption{Adversarial Example Generation}
	\label{alg1}
	
	\begin{algorithmic}[1]
		\scriptsize % Use smaller font size
		\Require $X \in \mathbb{R}^{N \times C}$ \Comment{N: number of points, C: dimensions}
		\Require $\tau (S,{\cal N})$ \Comment{the score of the sample's target class after subset $S$ is passed through the neural network ${\cal N}$ to extract features.}
		\Require $c1 \gets 0.25, c2 \gets 0.05$ \Comment{$c1$: Transformation scale parameter, $c2$: Perturbation parameter}
		\Require $r \in (0, 1)$ \Comment{$r$: selection ratio, fraction of points to be selected}
		\Ensure $X^{\text{adv}}$ $\in \mathbb{R}^{N \times C}$ \Comment{Adversarial sample}
		
		\Statex \textbf{\textcolor{red}{Shapley Value Calculation (Steps 1-9)}}
		\State $N, C \gets \text{shape(X)}$
		\State Initialize $\bm{\Phi}$ and set $\phi_i(\tau) = 0$ for each participant $i$. \Comment{Shapley values initialization}
		
		\For{each participant $i = 1$ to $N$}
		\For{each subset $S \subseteq \{1, 2, \dots, N\} \setminus \{i\}$}
		\State $|S| \gets$ number of members in subset $S$
		\State Compute contribution of participant $i$:
		\[
		\text{contribution} \gets {\frac{{(N - 1)! \cdot (|S| - 1) \cdot (N - |S|)!}}{{N!}}} \cdot (\tau (S,{\cal N}) - \tau (S\backslash \{i\},{\cal N}))
		\]
		\State Update the Shapley value:
		\[
		\phi_i(\tau) \gets \phi_i(\tau) + \text{contribution}
		\]
		\EndFor
		\EndFor
		
		\Statex \textbf{\textcolor{red}{Selection of Points Based on Shapley Values (Steps 10-12)}}
		\State $M \gets \lfloor r \times N \rfloor$ \Comment{Compute the number of points to select}
		\State $idx \gets \text{argsort}({\bm{\Phi}})[::-1]$ \Comment{Get indices of Shapley values in descending order}
		\State $X \gets X[idx[0:M]]$ \Comment{Select the top $M$ points based on Shapley values}
		
		\Statex \textbf{\textcolor{red}{Spatial Transformation (Steps 13-22)}}
		\State Initialize $k \in \mathbb{R}^{3 \times 3}$ with zeros \Comment{$k$: transformation matrix}
		\State $b, d, e, f \gets \text{random\_uniform}([c1 - 0.05, c1 + 0.05]) \times \text{random\_choice}([-1, 1])$
		\State $k[0, 0], k[0, 1], k[0, 2] \gets 1, 0, b$
		\State $k[1, 0], k[1, 1], k[1, 2] \gets 0, 1, d$
		\State $k[2, 0], k[2, 1], k[2, 2] \gets f, e, 1$
		\State $\delta \gets \text{random\_uniform}([-c2, c2], (N, C))$ \Comment{$\delta$: perturbation matrix}
		\State $X^{\text{adv}} \gets \text{matmul}(X, k) + \delta$ \Comment{\text{matmul}: matrix multiplication}
		
		\State \Return $X^{\text{adv}}$
	\end{algorithmic}
\end{algorithm}

\end{document}